**Automated Wildfire Damage Assessment from Multi-view Ground-level Imagery Via Vision Large Language Models**

*Miguel Esparza[1*], Archit Gupta[1], Kai Yin[2], Yiming Xiao[1], and Ali Mostafavi[1]*
*[1] Urban Reslience.AI Lab, Zachry Department of Civil and Environmental Engineering, Texas A&M University, College Station, TX 77840, United States of America*
*[2] Department of Computer Science and Engineering, Texas A&M University, College Station, TX 77840, United States of America*
**Corresponding Author: mte1224@tamu.edu*

## Abstract

The escalating intensity and frequency of wildfires demand innovative computational methods for rapid and accurate property damage assessment. Traditional methods are often time-consuming, while modern computer vision approaches typically require extensive labeled datasets, hindering immediate post-disaster deployment. This research introduces a novel, zero-shot framework leveraging pre-trained multimodal large language models (MLLMs) to classify damage from ground-level imagery. Using Generative Pre-trained Transformer 4o (GPT-4o) as the primary model with comparative validation against Qwen2.5-Vision-Language-32-Billion-Instruct (Qwen), the research evaluates two pipelines applied to the 2025 Eaton and Palisades fires in California. These pipelines include an end-to-end inference method (Pipeline A) and a decoupled workflow where visual cues drive text-based classification (Pipeline B). A primary contribution of this study is demonstrating the efficacy of MLLMs in synthesizing information from multiple perspectives. The findings show that while single-view assessments struggle to classify intermediate damage, a multi-view analysis yields dramatic improvements. To explore the impact of prompting methods, the research benchmarked a baseline zero-shot and heuristic approach against advance reasoning strategies (Structured-Chain-of-Thought and Self-Consistency). The results indicate that simple prompting methods achieve a comparable accuracy to the reasoning strategies. This parity provides practitioners the flexibility to tailor the workflow to their specific use case, prioritizing either computational efficiency or detailed reasoning without compromising performance. Overall, the practical contribution of this study is an immediately deployable, flexible, and interpretable workflow that bypasses the need for supervised training, significantly accelerating triage and prioritization for disaster response practitioners.

## 1. Introduction

The trajectory of recent wildfire seasons is alarming, evolving from natural ecological events into uncontrollable urban conflagrations that devastate the built environment (Dennison et al., 2014;

Mahmoud & Chulahwat, 2018). Events such as the 2023 Lahaina disaster—the deadliest in Hawaii's history, claiming 102 lives and inflicting $5.5 billion in damages (Insurance Institute for Business and Home Safety, 2024) —and Canada's unprecedented 2023 season, which burned 18.5 million hectares (Erni et al., 2024), underscore this escalating crisis. This threat is further exemplified by the January 2025 Eaton and Palisades fires in Los Angeles County, which destroyed roughly 16,000 structures. As urbanization expands deeper into the Wildland–Urban Interface (WUI) (Mahmoud, 2024; Taccaliti et al., 2023), the frequency and scale of destruction are increasing globally (Ahmed et al., 2018; Chas-Amil et al., 2012; K. Chen & McAneney, 2004; Salvati, 2014; Sarricolea et al., 2020). Yet the critical process of post-fire damage assessment—essential for allocating aid and initiating recovery—remains a significant bottleneck.

Post-fire damage assessment provides information critical to priority allocation of proper resources to those most in need and provides appropriate aid to the local community. In the United States, governmental agencies charged with disaster response, such as the Federal Emergency Management Agency (FEMA), are responsible for such damage assessments (Luo & Lian, 2024). In current practice, a preliminary damage assessment (PDA) is conducted by a group of trained individuals to inspect the physical damage of the affected residential homes, despite accessibility issues with buildings or residences. There are also concerns about the inspectors' safety. Moreover, the volume of damaged homes could overwhelm the PDA teams (Luo & Lian, 2024). The use of unmanned aerial vehicles (drones) to gather aerial imagery has been integrated into the PDA process. To process these images, computer vision models are developed but are used by FEMA mainly for prioritizing the deployment of in-person damage assessors, rather than being used for damage assessments (Luo & Lian, 2024). Moreover, aerial imagery for damage assessment has several limitations. Structures may be obstructed by clouds or smoke, especially in wildfire situations. Processing the images requires significant computational and storage capabilities. Finally, aerial imagery often lacks the resolution required to assess individual structures. These limitations motivate this research to explore the potential of multimodal large language models (MLLMs) and large language models (LLMs), for rapid damage assessment with ground-level images (GLIs).

The current landscape of post-wildfire PDA is constrained by significant operational bottlenecks. While remote sensing and aerial imagery provide high-level overviews, they frequently fail to deliver the granular detail required for accurate, structure-level assessments due to obstructions like smoke or vegetation canopy. Consequently, the reliance on manual, in-person inspections persists—an approach that is resource-intensive, slow, and exposes personnel to hazardous environments. Although recent research has GLIs as an alternative, the prevailing methodologies utilize supervised deep learning models, such as convolutional neural networks (CNNs) or vision image transformers (ViTs) (Luo & Lian, 2024; Yang et al., 2025). These models are inherently data-hungry, demanding large, labeled datasets that are impractical to generate rapidly in the chaotic aftermath of a disaster, thus failing to accelerate the assessment timeline when speed is most critical.

To bridge this gap, the objective of this research is to directly address the critical need for rapid and granular post-wildfire damage assessment, circumventing the limitations of manual inspections and the data dependency of traditional computer vision models. We propose a novel, zero-shot workflow utilizing pre-trained MLLMs to analyze GLIs. The study specifically targets a significant gap in the literature: the persistent inaccuracy in classifying nuanced damage (the *affected* category). To overcome this, our methodology is designed to synthesize multi-view perspectives, providing a holistic assessment beyond the front façade. Furthermore, the research systematically evaluates the efficacy of explicit domain knowledge by comparing a direct zero-shot MLLM approach (Pipeline A) against a two-stage, indicator-guided workflow where extracted visual cues drive text-based classification (Pipeline B). The rationale is to establish an accurate, interpretable, and immediately deployable framework that significantly accelerates the preliminary damage assessment process.

The remainder of this paper introduces a training-free, multi-view framework that synergizes MLLMs and LLMs for rapid parcel-level damage assessment using GLIs. Unlike existing approaches that rely on satellite data or require extensive annotated datasets and complex preprocessing for supervised learning, the proposed method can be deployed immediately in the aftermath of a wildfire event. The system leverages a pre-trained MLLM alongside various prompting methods to evaluate the model's damage indication capabilities. The research employs a phased evaluation strategy: first establishing a performance baseline by comparing Pipeline A and Pipeline B across both single-view (front façade) and multi-view configurations. In this baseline phase, a zero-shot prompting approach is utilized for Pipeline A, while a rigid, heuristic-based prompting approach is applied in Pipeline B. Once this baseline is established, the research explores the impact of advanced reasoning strategies such as Self-Consistency (SC) and Structured-Chain-of-Thought (S-CoT) on both pipelines. The core innovation of this work lies in the formalization of an interpretable, training free framework that decouples visual perception from logical reasoning, achieving high-fidelity classification without the computational costs of model training. The framework is validated using two 500-sample case studies from the Eaton and Palisades fires of January 2025, drawing data from the California Natural Resources Agency Damage Inspection (DINS) records. The methodology can be readily adapted to other natural hazards by modifying the indicator set. The results show that our approach achieves substantial performance improvements in the most challenging assessment category—the *affected* class—by effectively synthesizing multi-view evidence into consistent, reliable classifications across different events and pipeline configurations. From an operational perspective, this methodology addresses fundamental limitations of aerial imagery, including occlusion and insufficient parcel-level detail, while acknowledging that preliminary damage assessments remain resource-intensive. The framework provides emergency management practitioners with an immediately deployable, interpretable, and adaptable tool that can significantly accelerate post-fire triage and optimize resource allocation decisions.

## 1.1 Literature review

The literature review below is organized as follows: Section 1.1.1 examines how textual modalities, specifically LLMs, are used to mitigate natural hazards. Section 1.1.2 assesses the application of satellite imagery methods for wildfire damage assessment. Then section 1.1.3 examines the value of adding GLI data to damage assessment in both a flood and wildfire domain. Section 1.1.4 examines the application of computer vision language models.

### 1.1.1 Applications of LLMs for mitigating natural hazards

When a community is preparing for, responding, or recovering from a natural hazard situational awareness is critical (Fan et al., 2020). Natural language processing (NLP) has emerged as an important tool to enhance the dissemination of life-saving information and response coordination; however, traditional rule-based NLP approaches demonstrate limitations in managing the complexity of unstructured information related to natural hazards (Ma et al., 2022). To this end, LLMs present an opportunity to mitigate this limitation by leveraging deep semantic understanding to interpret complex, unstructured data streams without the rigidity of predefined rules. The technological advancement of LLMs was propelled by the transformer architecture (Vaswani, 2017), which allows the model to generate context-aware responses through self-attention mechanisms and parallel processing. The potential of these models has gained institutional and global recognition, as governmental agencies and policies are integrating the use of artificial intelligence (AI) technology into disaster management frameworks (Xu et al., 2025).

To evaluate the potential of LLMs for social sensing, McDaniel et al. (2024) explored the zero-shot capabilities of models like Generative Pre-trained Transformer 4o (GPT-4o) and Claude 3.5 in classifying crisis tweets (McDaniel et al., 2024). They found that while these models excel at binary informativeness tasks, they struggle with nuanced humanitarian categorization without specific event context. Addressing this limitation, Yin et al. (2024) introduced CrisisSense-LLM, an open-source model instruction-fine-tuned on over 60,000 tweets (Yin et al., 2024). This model enables simultaneous multi-label classification—identifying event types, informativeness, and aid needs—achieving a breakthrough accuracy of 63.8% and far surpassing traditional baselines. Liu et al. (2025) used retrieval-augmented generation (RAG) to establish a text-to-SQL benchmark that can enhance flood management decision making due to the multi-table queries with geospatial (Liu et al., 2025) reasoning. For wildfire management, Xie et al. (2025) developed WildfireGPT, an LLM agent integrated with RAG (Y. Xie et al., 2024). This system synthesizes location-specific climate data and scientific literature to provide expert-level risk analysis and mitigation strategies. Complementing this, Gao et al. (2025) introduced an "Instructor-Worker" multi-agent architecture that interfaces with Google Maps Platform's application programming interface (API) (K. Gao et al., 2025). This system autonomously retrieves and analyzes air quality data to generate valid health policy recommendations, demonstrating the viability of LLMs in automated environmental monitoring. Finally, Chen and Fang (2024) proposed a multi-round prompt engineering strategy that iteratively refines LLM responses, enhancing the accuracy and relevance of disaster issue consultations compared to standard tree of thoughts methods (W. Chen et al., 2024).

While textual modalities provide enormous potential for natural hazard mitigation, it is critical to explore multi-modal frameworks that integrate both text and vision. To this end, the following sections review the application of satellite imagery, ground-level imagery, and multimodal vision-language models in natural hazard settings.

### 1.1.2 Wildfire damage assessment via satellite imagery

Damage assessment is a crucial component of post-wildfire disaster management, providing insights into the severity and spatial extent of destruction. Traditional methods have relied on remote sensing and satellite imagery, as these technologies enable decision makers to prioritize interventions, allocate resources, and plan for long-term recovery. Wheeler and Karimi (2020) trained a Residual Network (ResNet) model on satellite imagery dataset that contained information on building damage after multiple natural hazards, such as a wildfire or flood (Wheeler & Karimi, 2020). Their model achieved an F1 score of 0.868. Galanis et al. (2021) used a ResNet18 model specifically to classify buildings as fully, partially, or non-damaged by training the model on post-fire satellite images (Galanis et al., 2021). The researchers trained a ResNet18 model on post-wildfire satellite images, achieving 92% and 98% accuracy on test sets for their damage classification. These models provided insight on proper resource allocation post-event but required large amounts of training data and manual labeling. To mitigate this limitation, post-event reports integrate geospatial layers, semantic inference, and socio-economic overlays to provide actionable insights. Xie et al. (2020) used satellite radar imagery for damage assessment by applying machine learning to radar images which can visualize damages through smokes and clouds (B. Xie et al., 2020). Lee et al. (2020) uses a semi-supervised workflow that allows their model to be trained on a small amount of labeled data with a large amount of unlabeled data and found this performance to be similar to methods that were fully supervised (Lee et al., 2020).

Despite efforts to mitigate the limitations of satellite imagery, damaged structures from these images may still be obstructed by smoke clouds or large vegetation, unlike ground level-images, thus hindering interpretation. Moreover, processing data for aerial images requires significant computing power and adequate data storage. Therefore, computation-efficient methods are critical to serve the need for tools that can process complex datasets for rapid damage assessment. For these reasons, there is a shift in this research field that endeavors to better understand how non-satellite imagery data could be utilized in the context of damage assessment.

### 1.1.3 Ground-level images for damage assessment

GLIs can gather more information about household vulnerability than satellite imagery, improving damage assessment for researchers, practitioners, and city managers. For example, Nia and Mori (2017) used a CNN-based model trained on a small manually curated GLI dataset to categorize buildings (ranging from no damage to severe damage). Their model achieved a 90% accuracy for the Ecuador and Nepal earthquake, as well as a 76% accuracy for Hurricanes Ruby and Matthew (Nia & Mori, 2017). Zhai and Peng (2020) applied a CNN to Google Street View (GSV) images to assess ground-level damage after Hurricane Michael struck Mexico Beach in Florida. Their model provided better insights compared to remote sensing images on household vulnerability by capturing damage to exterior walls, windows doors and facades (Zhai & Peng, 2020). Luo and Lian (2024) provided a ViT model to improve damage classification in a wildfire domain. They applied transfer learning to train a ViT model on roughly 18,000 ground-level images of homes. These homes had damage severity labels assessed by damage inspectors during the 2020–2022 California wildfires. Their model achieved an F1 score of 0.73 for *affected* households, 0.99 for *destroyed* households, and 0.95 for *no damage* households (Luo & Lian, 2024).

GLIs also help assess what household characteristics could make the property prone to a natural hazard. In the flood domain, the lowest floor elevation (LFE) is a key characteristic that can be determined from GLIs to better inform practitioners on the vulnerability of a household. Gao et al. (2023) used a deep learning computer vision model, You Only Look Once version-5 (YOLO-v5), to detect front doors to calculate the LFE of a household from GSV imagery by using a rectangular bounding box (G. Gao et al., 2024). Similarly, Ho et al. (2024) proposed a computer vision model, ELEV-VISION which uses a different bounding box (Ho et al., 2024). These methods allowed researchers to better detect front doors and the height difference from the street and lowest floor. These household elevation data, validated from ground truth elevation data by (Diaz et al., 2022) showed no statistical difference. To demonstrate the potential of GSV imagery to improve flood predictions, Esparza et al. (2025) used the household elevations from ELEV-VISION in flood depth-damage curves. They found statistical significance with both flood depth data and insurance claims (Esparza, Ho, et al., 2025). These works show the potential that ground level imagery has to be used in damage assessment by discovering key household features that could cause vulnerability.

To advance the potential that GLIs have in rapid damage assessment, the methodological contribution of the research is integrating these images into a multimodal vision and text-based framework. The current literature of these multimodal models is explored in section 1.1.4.

#### 1.1.4 Vision language models for post disaster assessment

The applications of vision language models in natural hazard research have opened opportunities for damage assessment; however, these models in damage assessment have been historically constrained by the scarcity of large-scale, annotated image-text datasets. Several studies to overcome this limitation by developing novel datasets and training frameworks. For instance, Wang et al. (2023) developed a large-scale remote sensing vision-language dataset using Google Earth imagery and OpenStreetMap data. Models trained on their dataset demonstrated an increased accuracy of 6.2% in zero-shot scene classification compared to models that are not trained on their dataset (W. Wang et al., 2023). Mall et al. (2024) introduced a framework that leverages paired satellite and ground-level imagery to train MLLMs without explicit textual annotations, showing that this approach outperforms traditional baseline models (Mall et al., 2024). Additionally, Wang et al. (2023) curated a multi-modal dataset for natural hazard analysis, encompassing 36 events related to floods, wildfires, and earthquakes (J. Wang et al., 2023). This specialized dataset not only aids emergency response planning but also improved model accuracy by 10.4% over baselines.

Multimodal vision and text models have also been used in damage assessment. For example, Ho et al. (2024) built upon their previous work of the ELEV-VISION model by adding text-prompted image segmentation, improving their previous results by obtaining a 75.63 Intersection over union segmentation task. The integration significantly enhanced the availability of LFE estimation from 33% of properties to 56%, covering nearly all houses (98.71%) where the front door was visible, while maintaining comparable accuracy to baseline models (Ho et al., 2024). Building on the utility of street-view imagery, Xiao et al. (2025) proposed Recov-Vision, a framework that links panoramic video to parcels for post-flooding occupancy detection; their two-stage strategy separating visual perception from reasoning significantly improved recall and recovery tracking compared to single-stage baselines (Xiao et al., 2025). Beyond flood scenarios, Estêvão (2024) demonstrated the feasibility of using generative AI models like GPT-4o to classify post-earthquake structural damage according to the European Macroseismic Scale 1998 (EMS-98), achieving notable accuracy in masonry and reinforced concrete structures (Estêvão, 2024). To address the specific challenges of wildfire smoke detection, Wu et al. (2025) introduced a modified Contrastive Language-Image Pre-training (CLIP) model, FireCLIP, to utilize a prompt tuning strategy which mitigates false positives from pseudo-smoke and adapt to regional data imbalances, improving zero-shot performance by over 12%. Similarly, Du et al. (2025) developed Qwen2-Wildfire by fine-tuning a lightweight multimodal model with optimized prompts, enabling detailed wildfire scene descriptions and accurate recognition (Du et al., 2025). Boroujeni et al. (2025) established a comprehensive review on how vision language and large language models can improve fire imagery analysis (Boroujeni et al., 2025). Moreover, Zhang et al. (2022) demonstrated the efficacy of combining visual and textual data through utilizing both a Bidirectional Encoder Representation from Transformers (BERT) and Visual Geometry Group (VGG-16) model, which achieved a 93% accuracy in classifying topics during the Henan rainstorm, significantly outperforming text-only baselines. Mendieta and Wen (2023) developed a multimodal model that processed both images (pre-fire aerial imagery) and text-based tabular data (household characteristics, weather features, and fire hazard rating). The tabular data was converted into descriptive text prompts and assessed in language models, such as Robustly Optimized BERT Pretraining Approach (RoBERTa) and GPT-2. The image dataset was processed with a ViT. This multimodal approach achieved an F1-score of 0.77 which outperformed models trained only on vision or text (Iván Higuera-Mendieta, Jeff Wen, 2023). Sun et al. (2023) developed a zero-shot vision question-answer model for flood scenarios that utilizes Chain-of-Thought (CoT) demonstrations to unlock reasoning capabilities, significantly improving answers to complex visual questions without pre-training (Sun et al., 2023).

The novelty of this research is to establish a methodological framework for practitioners to implement after a wildfire event to significantly enhance damage assessment. This is accomplished by using pre-trained MLLM and LLM tools to classify the damage extent of households based on GLIs.

## 1.2 Study objective

This study introduces two distinct pipelines to investigate how language models can enhance wildfire damage classification. There is a noticeable gap in the literature as current models struggle to properly classify households where the damage is not in the front façade (Luo & Lian, 2024). The proposed approach overcomes this by leveraging the capacity of MLLMs for semantic synthesis, treating each view of the building as an independent data point to construct a holistic damage profile that captures potential damage indicators hidden from a single viewpoint. Pipeline A and Pipeline B both process the frontal and comprehensive multi-view imagery. This study is driven by two objectives. First, the research aims to quantify the multi-view gain, determining how significantly the synthesis of multiple perspectives improves classification over single-view baselines. Second, the research interrogates the utility of prompting strategies, comparing a zero-shot (Prompt A) and rigid heuristics (Prompt B) against reasoning methods such as S-CoT (Prompt C) and SC (Prompt D). This comparison seeks to identify the optimal balance between high-fidelity metrics (F1 scores, AUC, and Cohan Kappa) and the operational value of auditable, reasoning-based outputs. These objectives ultimately aim to equip practitioners with a scalable, interpretable workflow that enhances both the speed and reliability of PDA.

# 2. Methodology

Section 2.1 will examine the study area and data processing. Section 2.2 will explain the construction of the proposed pipelines, computational resources, and prompting methods. Section 2.3 will provide an overview of the McNemar test and classification metrics this research uses.

## 2.1 Study area and data

During the month of January 2025, Los Angeles County experienced two catastrophic firestorms, the Palisades and Eaton fires. These fires evolved from wildland fires into uncontrollable urban fires and destroyed roughly 16,000 structures as well as claimed 30 lives. The conflagrations of these fires made both events the second (Eaton) and third (Palisades) most destructive fires in California's history. This was due to a combination of extreme weather conditions—such as droughts—before the fire; dense community construction (separation of structures were as low as 8 to 20 feet); and the on-going urban development in wildland areas (Insurance Institute for Business and Home Safety, 2025).

The Palisades and Eaton damage data comes from the California Natural Resource Agency Damage Inspection Program (DINS) database. The database contains records of wildland fires dating back to 2013 and documents the damage of structures in affected areas. The Palisades fire impacted roughly 12,000 structures, and the Eaton fire impacted roughly 18,000. The structures' damage classifications are as follows: *no damage*, *minor*, *affected*, *major*, and *destroyed*.

This study systematically consolidates the *minor*, *affected*, and *major* categories into a single *affected* class. While maintaining high granularity is desirable for policy and insurance contexts, recent literature demonstrates that the severe class imbalance inherent to these intermediate categories frequently degrades classification reliability. For instance, Zhang et al. (2025) observed in tornado assessments that differentiating between adjacent damage states often results in statistical noise rather than actionable insight (Zhang et al., 2025). Similarly, Noumeur and Tohir (2025) as well as Luo and Lian (2024) successfully employed similar aggregation strategies to improve predictive stability in wildfire modeling (Luo & Lian, 2024; Noumeur & Tohir, 2025). Supporting the previous literature, the cosine similarity results in table 2 and visual analysis in figures 3 and 4 reveal a high semantic overlap between the visual features of these intermediate classes; thus, this aggregation is implemented as a necessary assumption in the study to properly accomplish the research objectives. Future studies can explore the fine-tuning of open-sourced models to better capture granular damage, thereby mitigating the aggregation limitation inherent in this study.

This research uses the GPT-4o model to obtain the results in section 3. These results are compared to another open-sourced model, Qwen2.5-Vision-Language-32-Billion-Instruct (Qwen), where the same methodology is applied. This comparison is included in the discussion, and the respective tables that are used for the comparison can be seen in the supplementary material section (*S4*).

Pre-processing began with calculating the cosine similarity score with OpenAI's CLIP model for the entire dataset of Palisades and Eaton. The model will use an encoder to project the image inputs into an embedded vector space. Then, the cosine of the feature vectors are computed. This output is a measure of similarity between images based on high-dimensional feature representation. The equation for the cosine similarity score is shown below.

$$\textit{Cosine Similarity}\ (x, y) = \frac{x \cdot y}{||x|| \cdot ||y||} \quad (1)$$

In the above equation, $x$ is the *affected* category while $y$ would be either *minor* or *major* category. The dot product of the vectorized images is taken and it is divided by the product of the magnitude of these categories. A higher score indicates that the categories show similarities in these images.

An additional pre-processing step involved taking 500 random building samples for each study area. To explore the extent to which the number of images impact the classification results, refer to supplementary material section *S1*. Since the aggregation was feasible for the whole dataset, no cosine similarity scores were computed on the samples. Taking a smaller sample set was necessary due to computational and monetary limitations of running the pre-trained model. Additionally, the results would only marginally improve since the model is pre-trained despite more computational and monetary resources. Figure 1 shows the study areas and sample data for Palisades and Eaton fires.

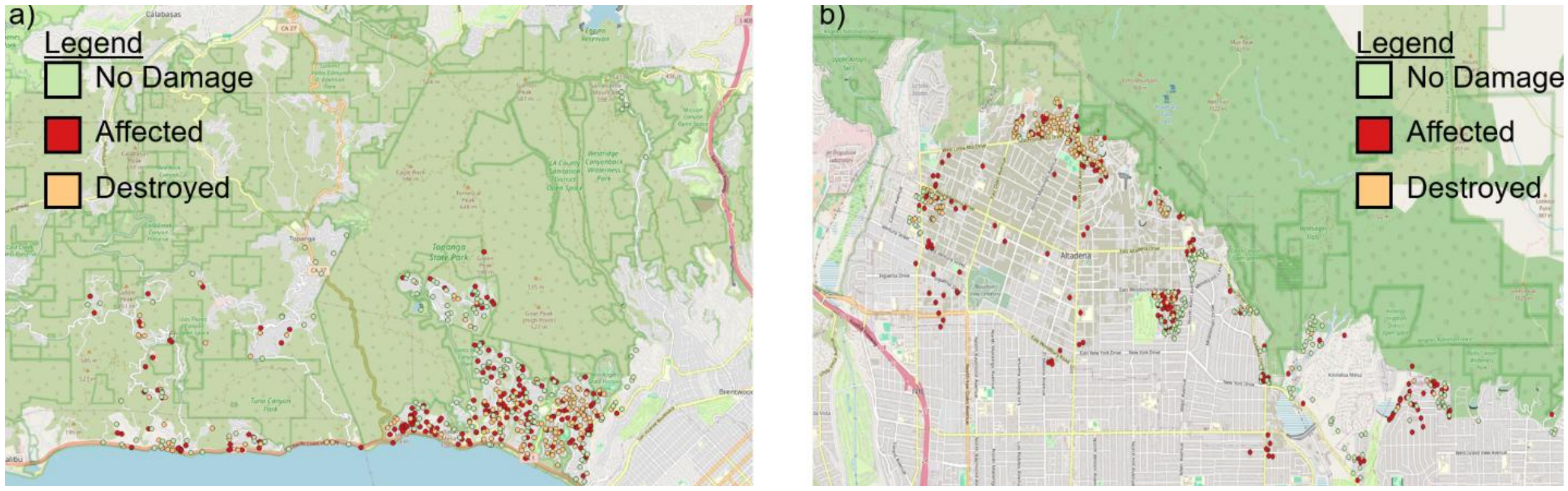

**Figure 1: Study area of 500 building samples in (a) Palisades Fire and (b) Eaton Fire.**

## 2.2 Methodological Framework

This study establishes a comprehensive workflow for holistic wildfire damage assessment synthesized through the capabilities of MLLMs. The system processes lists of base64-encoded images in a unified sequence, allowing for the integration of multiple perspectives per household. The research evaluates the efficacy of this approach through Pipeline A and Pipeline B which use the GPT-4o model, accessed via the OpenAI chat completion's API. To ensure experimental reproducibility, no specific model snapshot was pinned; however, hyperparameter controls were applied as seen in table 3. Preprocessing involved sequential image loading and base64 encoding without local graphics processing unit (GPU) acceleration.

Figure 2 shows the workflow of both proposed pipelines. Pipeline A only uses the MLLM model, after the images (which are not augmented) are passed in base64 format. Then the pipeline is tasked to classify the images as *no damage*, *affected*, or *destroyed* based on the selected prompting method. Pipeline B is a modified version of Pipeline A, where the MLLM outputs visual cues, in a JSON format, that identify damage indicators from the encoded images. These outputs from the MLLM will either be binary (Prompt B) or its reasoning logic (Prompt D). Based on the prompt selection for Pipeline B, the LLM engine will make its classification. Both pipelines are used to analyze a single front view image of a household, and the other perspective of a house when multiple images are available. Section 2.2.1 outlines the application of the different prompting methods. Section 2.2.2 and section 2.2.3 outline the details of Pipeline A and Pipeline B respectively.

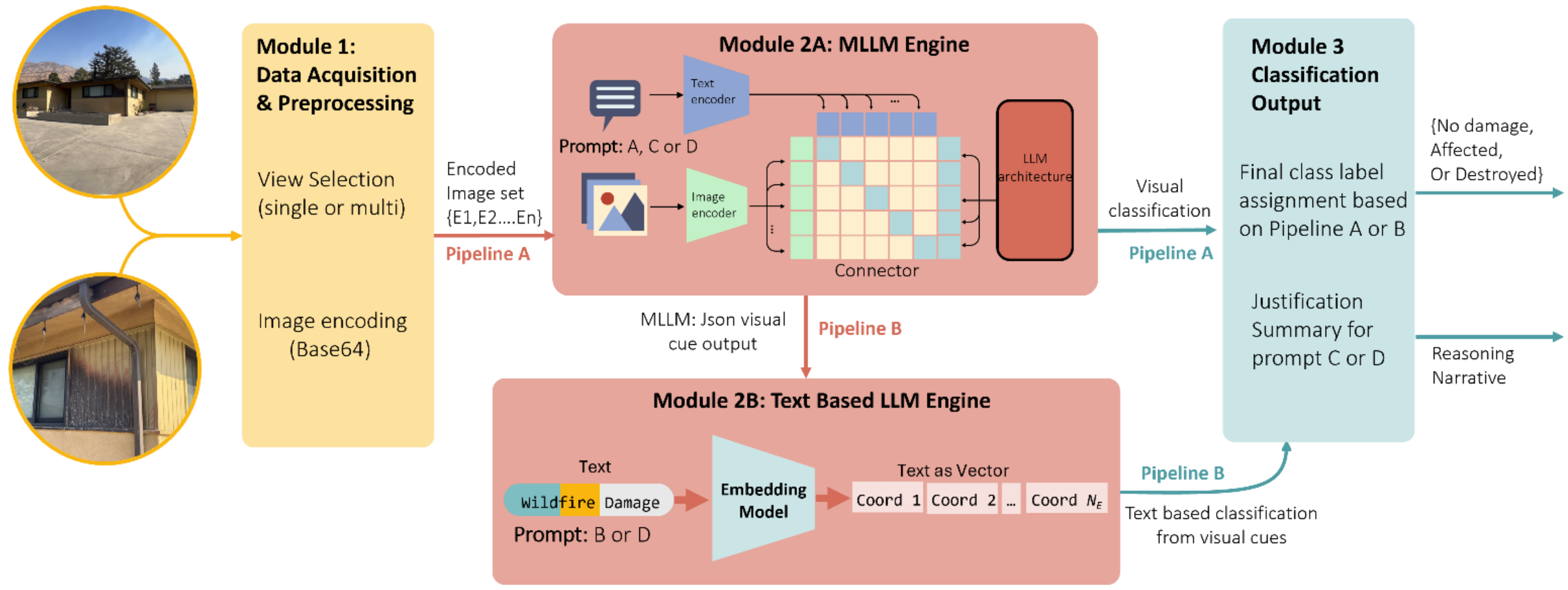

**Figure 2: Module Diagram of the Pipeline Framework. Module 1: The single-view or multi-view imagery is base64-encoded and processed with the model's internal vision encoder. Module 2A: The encoded visual embeddings are passed through a connector to project them into the text embedding space, enabling concatenation with the specified prompt instructions (Prompt A: zero-shot, Prompt C: Self-Consistency, Prompt D: Structured-Chain-of-Thought). The combined sequence is processed by the open-source's LLM architecture backbone to generate the damage classification (Module 3). Module 2B: This decoupled augmentation is a two-stage workflow where the MLLM first extracts the specific damage indicators (as boolean values or structured reasoning traces), which are fed into a text-based model to determine the final classification (Module 3) based on a rigid list (Prompt B) or reasoning approach (Prompt D).**

### 2.2.1. An overview of the applied prompting methods

Prompting is a set of instructions given to the MLLM and LLM to generate the classification output based on the input image (Estêvão, 2024). Since prompting is a key driving force in the model's decision making, altering the strategies can influence the results (Ji et al., 2023).

Therefore, to evaluate the efficacy of MLMMs in damage assessment, this research evaluates four prompting strategies: (1) zero-shot, (2) heuristic attributed-based, (3) Self-Consistency (SC), and (4) Structure-Chain-of-Thought (S-CoT). These strategies were applied across Pipeline A and Pipeline B to isolate the impact of visual perception versus reasoning logic. Prompt B, Prompt C, and Prompt D were developed based on domain indicators found in literature related to individual building damage assessment. Meldrum et al. (2022) provides a parcel level risk framework that describes how wildfire risk varies from home to home based on structural features, fire characteristics, and how cosmetic damage could indicate the impact of a fire (Meldrum et al., 2022). Chulahwat et al. (2022) and Esparza et al. (2025) explored the impact of environmental and topological features can impact the spread of a fire (Chulahwat et al., 2022; Esparza, Battal, et al., 2025). Therefore, the aforementioned three prompts focus on features related to damage to the building envelope, windows, burn marks on the building or surrounding environment, and nearby vegetation. The exact prompting text can be seen in the respective algorithm pseudo code. The explanation of the prompt strategy and applied pipeline is seen below:

*Zero-shot (Prompt A)*: Applied to Pipeline A, this approach establishes a baseline for the model's innate capability to classify damage without domain-specific guidance. It tests the model's pre-trained knowledge against the specialized task of damage assessment.

*Heuristic Attribute-Based (Prompt B)*: Applied to Pipeline B, this method introduces a structured check list of domain-specific damage indicators. The model extracts visual cues from the MLLM, which serves as the rigid input for the text-based classifier LLM.

*Self-Consistency (Prompt C):* Implemented in Pipeline A, this strategy, inspired by (X. Wang et al., 2022), addresses the inherent stochasticity of LLMs. By re-sampling the visual evidence to generate k=5 independent reasoning paths, thus mitigating the risk of hallucinations. This method posits that while a model may hallucinate in a single pass, the consensus across diverse reasoning paths yields a more robust classification.

*Structured Chain-of-Thought (Prompt D):* Applied to both pipelines, this prompt adapts the Chain-of-Thought (CoT) methodology (Wei et al., 2022) by imposing a hierarchical logic structure: it forces the model to first attempt to validate a "pristine hypothesis" (Thought A) before considering a "damage hypothesis" (Thought B), and finally applying a "context filter" (Thought C) to ensure only nearby environmental features are focused, rather than background noise. The rationale for this approach, is that language models can have an enhancement in classification with hierarchical logic rather than open-ended reasoning (Boroujeni et al., 2025).

### 2.2.2 An overview of the vision-based wildfire damage classification MLLM pipeline

For Pipeline A, the image(s) are provided directly to the model. Under Prompt A, the pipeline classifies wildfire damage into the *no damage*, *affected*, or *destroyed* categories without additional context Thus, Pipeline A – Prompt A lacks an intermediate step for visual parameters. This gives the model ability to classify a house without abiding by the cues given to it; however, this simplicity also limits the extent to which data can be interpreted and the confidence of along with domain-specific analysis for these images. To mitigate this "black-box" limitation and enhance interpretability, the research introduces a secondary evaluation phase incorporating Prompt C and Prompt D. These prompts are applied after benchmarking the single-view and multi-view performance of the baseline Prompt A.

A key capability of this MLLM framework is its ability to synthesize multi-view data by processing sequences of encoded images. For households with multiple perspectives, the system aggregates the respective base64-encoded representations into a unified input list, treating the views as a cohesive context window. This allows for a direct comparative analysis between single-view and multi-view workflows to quantify the performance gain attributed to comprehensive visual coverage. The operational logic for this pipeline is formalized in Algorithm 1 and assumes the open-sourced model is GPT-4o and Prompt A (baseline configuration) for classification. Pipeline A – Prompt C and Pipeline A – Prompt D can be seen in Algorithm 3 and Algorithm 4, respectively, in supplementary material section *S2*.

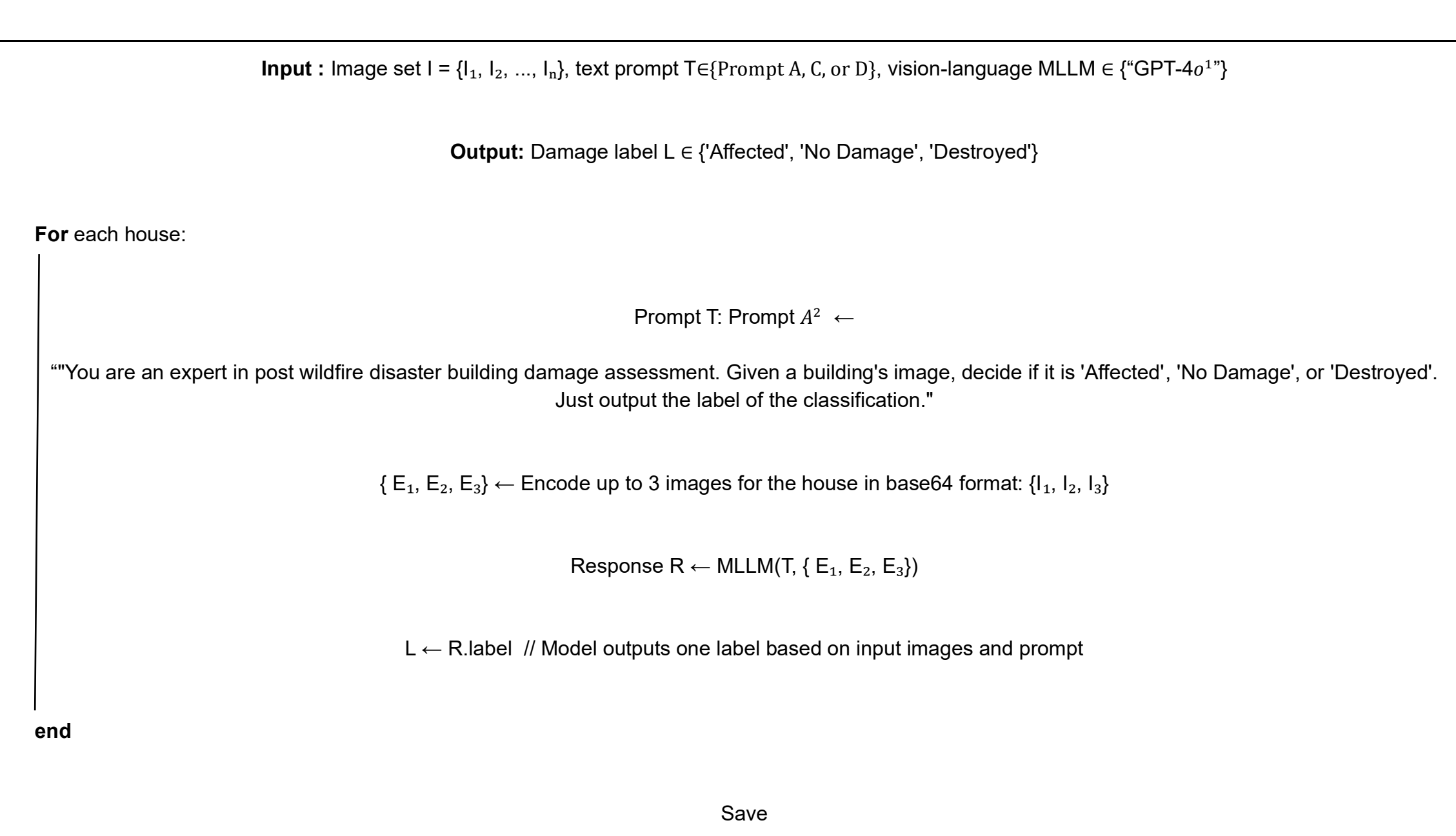
**ALGORITHM 1: Pipeline A – Prompt A**

**Input :** Image set I = {$I_1$, $I_2$, ..., $I_n$}, text prompt T∈{Prompt A, C, or D}, vision-language MLLM ∈ {“GPT-4$o^1$”}

**Output:** Damage label L ∈ {'Affected', 'No Damage', 'Destroyed'}

**For** each house:

Prompt T: Prompt $A^2$ ←

“"You are an expert in post wildfire disaster building damage assessment. Given a building's image, decide if it is 'Affected', 'No Damage', or 'Destroyed'. Just output the label of the classification."

{ $E_1$, $E_2$, $E_3$} ← Encode up to 3 images for the house in base64 format: {$I_1$, $I_2$, $I_3$}

Response R ← MLLM(T, { $E_1$, $E_2$, $E_3$})

L ← R.label // Model outputs one label based on input images and prompt

**end**

Save

1. *The application of this algorithm can be used with GPT-4o or Qwen*
2. *The application of Prompt C and Prompt D with respect to Algorithm 1 can be seen in the supplementary material section S2.*

### 2.2.3 An overview of the text-based wildfire damage classification decoupled pipeline

Pipeline B, built by augmenting Pipeline A, includes an intermediate output from the MLLM. For Prompt B, the MLLM processes the image(s) of the houses and outputs a set of structured indicators along with their corresponding true or false values. Then, the outputs are passed to an LLM, which integrates these cues with the contextual prompt to determine the label for the houses. Pipeline B is applied to both single-view and multi-view image tests. The results are first compared internally to address the first research objective. Once the first objective is achieved, the research examines the extent to which LLM prompting strategies can enhance damage assessment. Specifically with Pipeline B – Prompt D, the research observes the ability of LLMs to process abstract logic without visual cues. For Prompt D, the MLLM generates a natural language reasoning block instead of structured booleans, which the secondary LLM must interpret to derive the damage state. The operational logic for Pipeline B – Prompt B is *n*ormalized in Algorithm 2 and the application of Pipeline B – Prompt D can be seen in Algorithm 5 in the supplementary material section (S2).

**Algorithm 2: Pipeline B – Prompt B**

---

**Input:** Image set I = $\{I_1, I_2, ..., I_n\}$, structured text prompt $T_1 \in \{\text{Prompt B or D}\}$, MLLM ∈ {"GPT-4o"}, indicator-based prompt $T_2$, LLM ∈ {" GPT-4o[1]"}

**Output:** Damage label L ∈ {'Affected', 'No Damage', 'Destroyed'}

**For** each house:

MLLM Indicator prompt $T_1$: Prompt$B$[2] ←

"Analyze the image and answer with a JSON object in the following format:

{

'is the house destroyed': true/false,

'is the structure damaged': true/false,

'is the glass or windows broken': true/false,

'is the furniture burnt': true/false,

'are there burn marks on the structure': true/false,

'is the vegetation around burnt': true/false

}

Only output the JSON object, nothing else."

$\{E_1, E_2, E_3\}$ ← Encode up to 3 images for the house in base64 format: $\{I_1, I_2, I_3\}$

Indicators JSON V ← MLLM($T_1$, $\{E_1, E_2, E_3\}$)

LLM prompt $T_2$: Prompt $B$[2] ←

"You are an expert in post wildfire disaster building damage assessment.

Given a building's attributes in JSON, decide if it is 'Affected', 'No Damage', or 'Destroyed'.

If the attribute says destroyed is true, then output 'Destroyed'.

If any one of the other attributes are true, then output 'Affected'.

If none of the attributes are true, then output 'No Damage'.

Now, decide for this building:

{Indicators JSON V}

Output only one word: 'Affected', 'No Damage', or 'Destroyed'."

Response R ← LLM($T_2$)

L ← R.label // Model outputs one label based on indicators JSON

**end**

Save

---

1. *The application of this algorithm can be used with GPT-4o or Qwen*
2. *The application of Prompt D with respect to Algorithm 2 can be seen in the supplementary material section, S2.*

## 2.3 Statistical testing and classification parameters

This research uses precision (equation 2), recall (equation 3) and the F1 score (equation 4) score to evaluate the classification results of each pipeline.

$$Precision\ (P) = \frac{TP}{TP+FP} \quad (2)$$

$$Recall\ (R) = \frac{TP}{TP+FN} \quad (3)$$

$$F1\ score = 2 * \frac{P*R}{P+R} \quad (4)$$

These metrics are the standard criterion for evaluating classification problems (Yang et al., 2025). $TP$ refers to a true positive classification which is an image that was correctly classified into its respective damage level. $FP$ refers to a false positive and occurs when the model incorrectly assigns an image into a class. $FN$ is a false negative and occurs when the model overlooks class features of an image. The F1 score is calculated as the harmonic mean of $P$ and $R$ which provides a balanced metric to evaluate the model's classification performance. This metric will be used when discussing the results of the pipelines.

To extend the classification metrics, the research incorporates Cohan Kappa to quantify the inter-rater reliability between the model predictions and ground truth, correcting for agreement occurring by chance. The Cohan Kappa can be computed as seen in equation 5, where $p_o$ is the observed agreement and $p_e$ is the expected agreement by chance.

$$\kappa = \frac{p_o - p_e}{1-p_e} \quad (5)$$

Furthermore, the research examines the Receiver Operating Characteristic-Area Under the Curve (ROC-AUC). The ROC-AUC provides a threshold-independent measure of the model's discriminative capability, specifically evaluating how well the pipelines distinguish between the damage categories across different confidence thresholds.

To statistically compare the performance of different modeling approaches, we employed McNemar's test. This test is suitable for analyzing paired categorical data, allowing us to evaluate the marginal homogeneity between two models' predictions on the same dataset. A two-by-two contingency table was constructed for each comparison, tabulating the instances where both models were correct, both were incorrect, and where they disagreed. The test focuses specifically on the asymmetry of the disagreement counts. A statistically significant result ($p < 0.05$) indicates that the two models have different error rates. We conducted this test for three distinct cases:

Case 1: single-view vs. multi-view results within Pipeline A to evaluate the impact of a multi-view image assessment.

Case 2: single-view vs. multi-view assessment within Pipeline B to evaluate the impact of a multi-view image assessment.

Case 3: multi-view assessment of Pipeline A vs. Pipeline B to evaluate the impact of LLM prompting.

The McNemar's coefficient is calculated by equation 6, where $b$ and $c$ are the classification results for the pipelines.

$$\chi^2 = \frac{(|b-c|-1)^2}{b+c} \tag{6}$$

A high $\chi^2$ value allows the research to reject the null hypothesis. In this application, the null hypothesis for case 1 and case 2 would be that there is no statistical significant difference in the error rates between the single-view and multi-view assessment; therefore, both image assessments perform equally well. By rejecting the null hypothesis for these two cases, the potential improvement between a single-view and multi-view assessment does not randomly occur. For case 3, similar logic applies except, this will determine if any improvement that LLM prompting can provide for the multi-view image assessment occurs randomly.

## 3. Results

To examine the extent to which the proposed methodology can improve wildfire damage assessment, section 3.1 will contextualize the data aggregation with the cosine similarity score and a visual analysis. Additionally, this section overviews the computational resources this workflow requires. Section 3.2 presents the classification metrics of both pipelines using the baseline prompt configurations (Prompt A and Prompt B) on single front-view GLIs for the Palisades and Eaton datasets. Section 3.3 extends this analysis by applying the same configurations and prompts to multi-view imagery where available. In section 3.4, the classification results from the single-view and multi-view experiments are statistically compared using McNemar's test and analyzed via class distribution charts and ROC-AUC charts. To further support these statistical findings, section 3.5 provides a qualitative visual analysis comparing successful and failed cases between the single-view and multi-view approaches. To address the second research objective, section 3.6 compares the reasoning-based prompts against the baseline prompts, applied to the superior viewpoint configuration identified in the previous sections. Finally, section 3.7 offers a qualitative visual analysis to illustrate the reasoning logic and comparative performance of these prompting strategies.

### 3.1 Data contextualization

Table 1 compares the number of households in each category for the original dataset and the sampled dataset for both study areas. Few households in either the total or sample datasets in both study areas are classified as sustaining *minor* or *major* damage, implying an imbalanced dataset. This classification prompted the need for the research to aggregate the data into the *affected* category. To computationally validate this aggregation, the cosine similarity scores were calculated with OpenAI's CLIP model based on the encoded images.

**Table 1: Number of households in each damage category**

| Category | Eaton Total Data | Eaton Sample Data | Palisades Total Data | Palisades Sample Data |
|---|---|---|---|---|
| **No Damage** | 7,894 | 155 | 4,262 | 155 |
| **Affected** | 856 | 181 | 730 | 178 |
| **Minor** | 148 | 6 | 171 | 8 |
| **Major** | 70 | 3 | 72 | 4 |
| **Destroyed** | 9,413 | 155 | 6,835 | 155 |

Table 2 shows the cosine similarity score for both study areas. The *affected* versus *minor* (Eaton: 0.794; Palisades: 0.730) and *affected* versus *major* (Eaton: 0.766; Palisades: 0.728) categories are rather high and are comparable to the scores for within the remaining categories.

**Table 2: Cosine similarity score**

| Category | Eaton Cosine Score | Palisades Cosine Score |
|---|---|---|
| **Affected vs minor** | 0.794 | 0.730 |
| **Affected vs major** | 0.766 | 0.728 |
| **Within affected** | 0.797 | 0.730 |
| **Within destroyed** | 0.861 | 0.753 |
| **within no damage** | 0.797 | 0.727 |

To further justify the aggregation of intermediate damage categories, figure 3 and figure 4 present visualizations of specific damage features. Visually, the buildings in the *minor*, *affected*, and *major* categories remain structurally standing despite sustaining localized damage to features such as siding, roofs, or windows. It is evident that these three categories share a certain extent of visual similarity with each other, while differing immensely from the *destroyed* (figure 10) category. Consequently, the *minor*, *affected*, and *major* categories are grouped together similar to the methodology in Luo and Lian (2024). This decision is supported by three converging factors: the small sample size of the intermediate classes (table 1), the cosine similarity results (table 2), and the visual indistinctness between intermediate damage and the collapsed structures (*destroyed*).

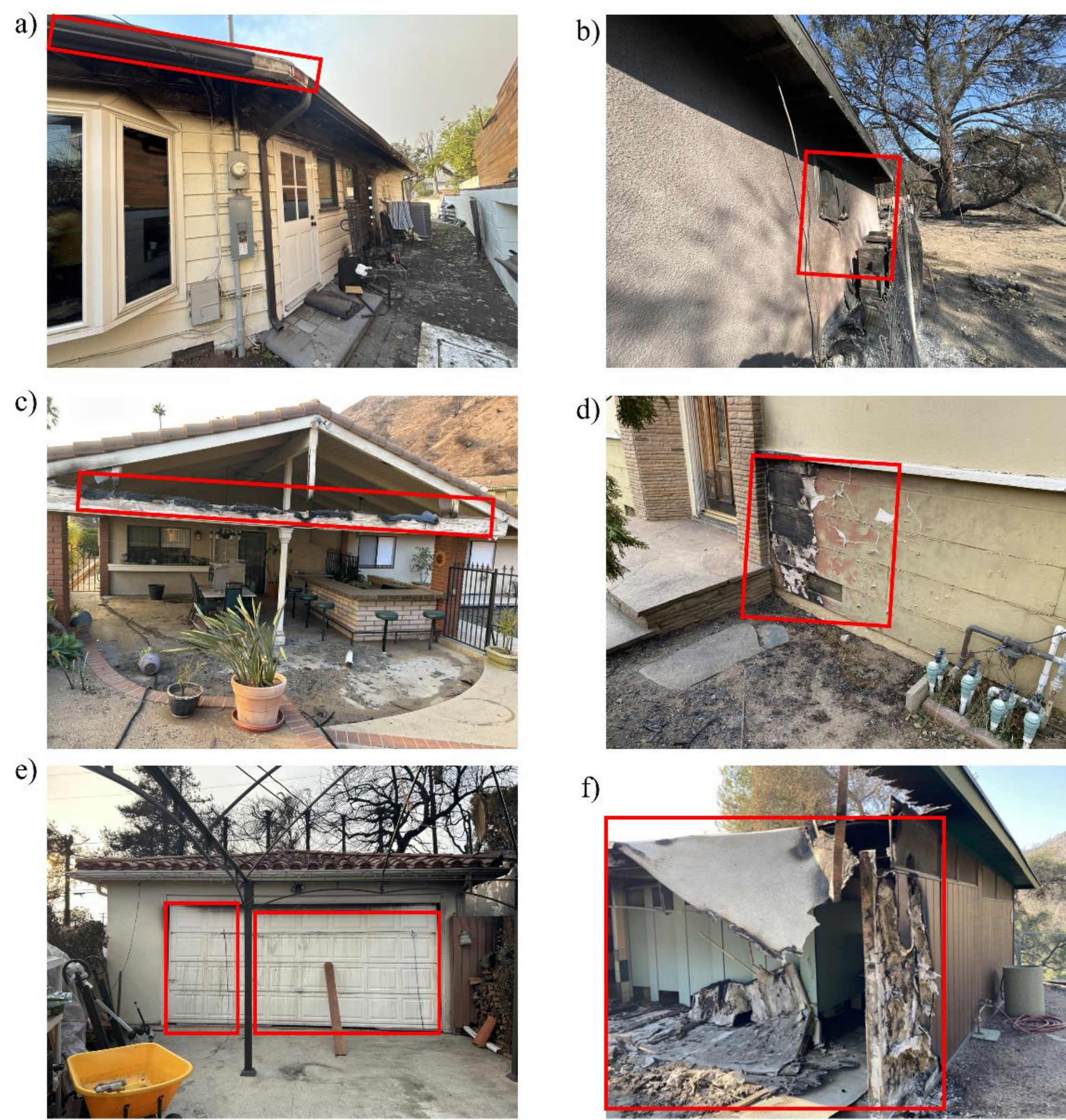

**Figure 3: Each image is an individual building impacted by the Eaton fire. 3a – 3b are *minor* damage buildings, 3c – 3d are *affected* buildings, and 3e - 3f are *major* damage buildings. All buildings have annotations of where the damage occurred.**

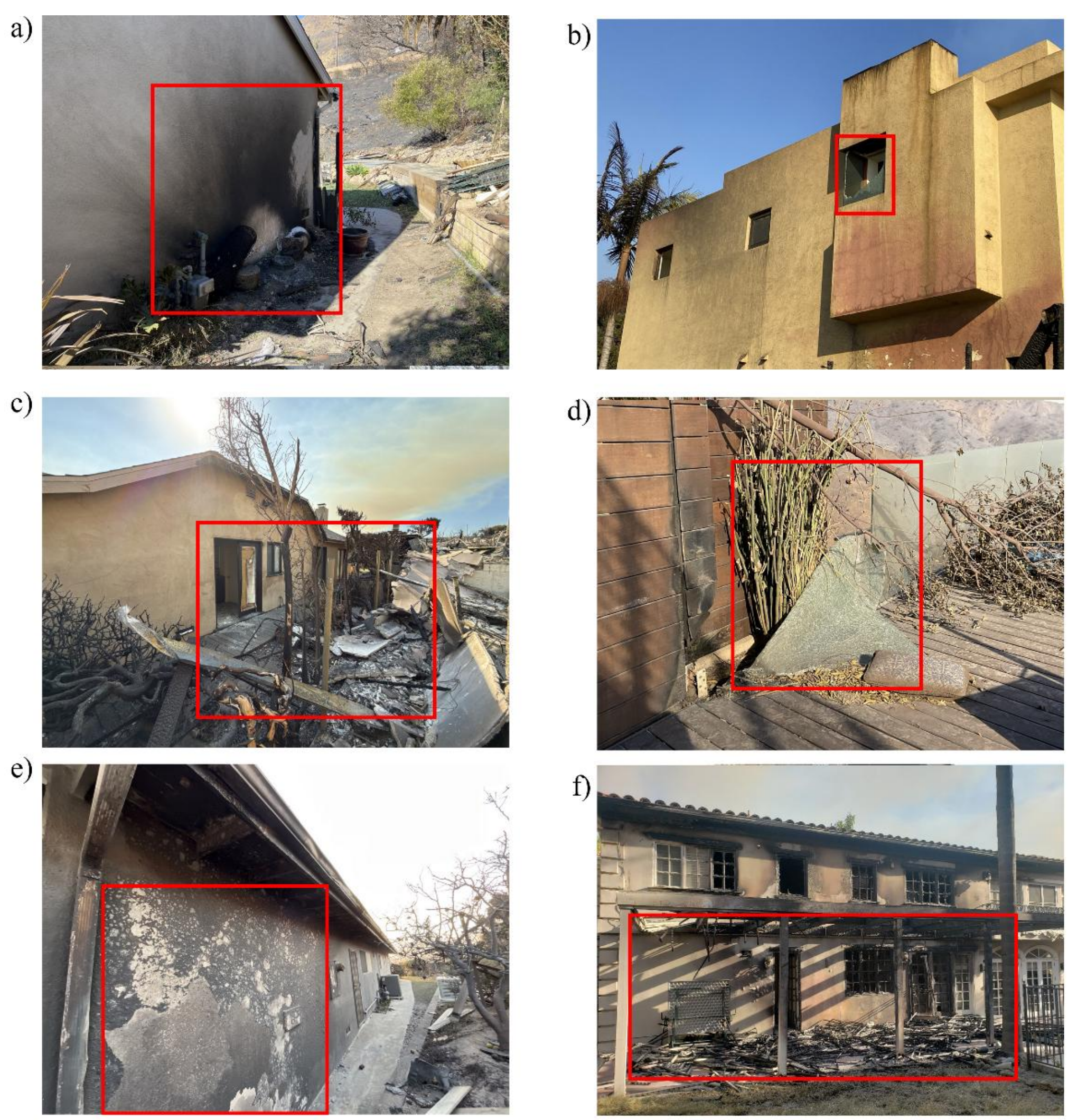

**Figure 4: Each image is an individual building impacted by the Palisades fire. 4a – 4b are *minor* damage buildings, 4c – 4d are *affected* buildings, and 4e - 4f are *major* damage buildings. All buildings have annotations of where the damage occurred.**

The research took 500 random samples because of monetary and computational constraints The usage of MLLM and LLM requires monetary resources based on tokenization, which is the process of transforming images into compressed representative units that can be processed by these models. The number of tokens used depends on the length and complexity of both the input and output of the model. Each experiment had 500 samples, with some samples containing more than one image. Whenever an API call is made to the model, both the input prompt in the form of text and images, and the output generated for the classification contribute towards the count of total tokens used. To manage token usage and ensure output consistency, the max tokens hyperparameter, which sets the maximum length of the generated output, and the temperature hyperparameter, which controls the randomness of the output, were controlled for each pipeline stage. Table 3 shows the temperature and max token hyperparameters for each pipeline stage with the respective prompt method. The max tokens parameter was matched to the expected response length: low for simple classifications, and high for structured outputs. Concurrently, the temperature was set low for the deterministic task and moderate for classification task.

**Table 3: Hyperparameters for the pipeline and prompt pairs**

| Hyper parameter | Pipeline A Prompt A | Pipeline A Prompt C | Pipeline A Prompt D | Pipeline B Prompt B | Pipeline B Prompt D |
|---|---|---|---|---|---|
| **Temperature** | 0.5 | 0.7 | 0.2 | [MLLM: 0.2; LLM: 0] | [MLLM: 0.2; LLM: 0] |
| **Max Token** | 5 | 800 | 1500 | [MLLM: 200; LLM: 5] | [MLLM: 1500; LLM: 5] |

Table 4 shows the computational resources required to implement the pipeline prompt pairs. In Pipeline A, the tokens are counted for the single stage process where the MLLM receives image(s) as input and directly outputs the classification. In Pipeline B, tokens are counted separately for both the stages (MLLM and LLM) where the MLLM receives image(s) along with a guided prompt and the LLM inputs the computed indicator values along with the context. This token counting is done for both the single and multiple images scenarios for each pipeline. For this purpose of cost calculation, we sampled 10 records under each experimental scenario and measured the input and output token count for each component in the pipelines. The operational cost for each pipeline was calculated based on per-token pricing, with input tokens at $ $2.5\times10^{-6}$ and output tokens at $ $1.0\times10^{-5}$ according to the OpenAI platform (OpenAI, 2024). For Pipeline A, the cost per sample varied by prompting strategy. The baseline (Prompt A) was the most economical, ranging from $0.0013 (single image) to $0.0039 (three images). The structured reasoning approach (Prompt D) incurred negligible additional expense, ranging from $0.0014 to $0.0039 per sample. However, the self-consistency approach (Prompt C) was significantly more expensive, ranging from $0.0069 to $0.0197 per sample due to the high token volume required for multiple inference paths. The prompts in Pipeline B added a consistent overhead due to the utilization of LLMs. This cost ranged from $0.0025 to $0.0051 for Prompt B and $0.0027 to $0.0053 for Prompt D. While these costs remain relatively low, making the method monetarily feasible for post-hazard deployment, users must balance the trade-off between the additional cost of integrated reasoning and implementing efficient methods such as zero-shot prompting.

**Table 4. Summary of the computational resources of the APIs for the MLLM and LLM**

| | **Pipeline A** | | | **Pipeline B** | | |
|---|---|---|---|---|---|---|
| Tokens per 10 samples | **Multiple images (2)** | **Multiple images (3)** | **Ground level images** | **Multiple images (2)** | **Multiple images (3)** | **Ground level images** |
| MLLM input tokens | [9,320; 46,825; 9,873] | [15,440; 77,425; 15,485] | [5,240; 26,425; 5285] | [9,770; 9,858] | [15,890; 15,988] | [5,690; 5,978] |
| MLLM output tokens | [14; 330; 66] | [15; 332; 68] | [18; 333; 70] | [660; 719] | [662; 723] | [608; 612] |
| LLM input tokens | [0;0;0] | [0;0;0] | [0;0;0] | [1,930; 2,140] | [1,930; 2,142] | [1,904; 2139] |
| LLM output tokens | [0;0;0] | [0;0;0] | [0;0;0] | [11; 50] | [12; 50] | [18; 52] |
| Total price for 1 sample ($) | [0.0023; 0.0120; 0.0025] | [0.0039; 0.0197; 0.0039] | [0.0013; 0.0069; 0.0014] | [0.0036; 0.0038] | [0.0051; 0.0053] | [0.0025; 0.0027] |
| Total price for 10 samples ($) | [0.0230; 0.1200; 0.0253] | [0.0387; 0.1969; 0.0394] | [0.0130; 0.0694; 0.0139] | [0.0360; 0.0377] | [0.0510; 0.0530] | [0.0250; 0.0270] |

For Pipeline A: [Prompt A; Prompt C; Prompt D] For Pipeline B: [Prompt B; Prompt D]

Table 5 delineates the trade-offs between architectural complexity and operational latency. The average latency was calculated by measuring the end-to-end runtime of 500 samples processed sequentially via the API. To estimate the energy consumption of the proposed framework, the research utilized the hardware power profiles established by (Luccioni et al., 2023). They ran a 176 billion parameter model for 432 hours and the total energy was 914 Kilowatts (kW)-hours; therefore, the average power draw for this large scale model was 2.11 kW. OpenAI has not publicly disclosed the parameter count for GPT-4o due to the competitive landscape of large models (Hurst et al., 2024). Therefore, the research provides a range of approximate energy values, which are in kilojoules (kJ), in table 5. The upper bound assumes the model operates at the full 2.11 kW power profile. The lower bound value, 0.96 kW applies a conservative utilization factor of roughly 45%, representing the estimated marginal dynamic power consumption. This accounts for situations where not all parameters are active or the efficiency gains of a mixture of experts architecture. The energy estimations were derived by multiplying the upper and lower power bounds by the latency values recorded for each pipeline.

Among the five experiments, Pipeline A – Prompt A is the most efficient in terms of time (2.2 s) and energy (2.1 – 4.6) kJ; however, it lacks the explicit reasoning capability critical for auditable damage assessment. Conversely, Pipeline A – Prompt C is significantly more resource-demanding, requiring approximately 20 seconds and (19.6 – 43.0) kJ per parcel due to the five-fold iteration required by the SC methodology. While the decoupled architecture of Pipeline B generally consumes more resources due to its multi-stage latency, it offers a unique operational advantage: if visual descriptions are cached, users can run the text classification stage independently. This text-only update consumes minimal resources (1.9 – 4.2) kJ, allowing for rapid re-classification. From a resource perspective, Pipeline A – Prompt D provides a balance, securing detailed reasoning with moderate resource efficiency (3.9 – 8.7 kJ). Together, tables 4 and 5 demonstrate the resource trade-offs inherent to each experiment, highlighting the workflow's flexibility to adapt to scenarios requiring zero-shot speed, decoupled text agility, or integrated reasoning.

**Table 5. Computational resources for each pipeline and prompt pair**

| Computational Resources | Pipeline A Prompt A | Pipeline A Prompt C | Pipeline A Prompt D | Pipeline B Prompt B | Pipeline B Prompt D | Text classification only |
| --- | --- | --- | --- | --- | --- | --- |
| **Average latency per image in seconds (s)** | 2.2 s | 20.4 | 4.1 s | 5.5 s | 6.1 s | 2 s |
| **Estimated energy per image in kJ (lower bound – upper bound)** | (2.1 – 4.6) kJ | (19.6– 43.0) kJ | (3.9 – 8.7) kJ | (5.3 – 11.6) kJ | (5.9–12.9) kJ | (1.9–4.2) kJ |

### 3.2 Application of language models in a single-view front image analysis for damage assessment

The single front view GLIs are assessed with both pipelines in each study area. Figure 5 shows the confusion matrix for Eaton and Palisades. In all four-confusion matrixes, the prediction for *destroyed* (97.4 % to 100%) and *no damage* (92.3% to 97.4%) were generally classified correctly; however, the *affected* category (13.2% to 36.3%) had the most misclassifications, as most of these images were classified as *no damage*. This shows that using a single image does not capture the full extent of damage since the building could be impacted at other locations.

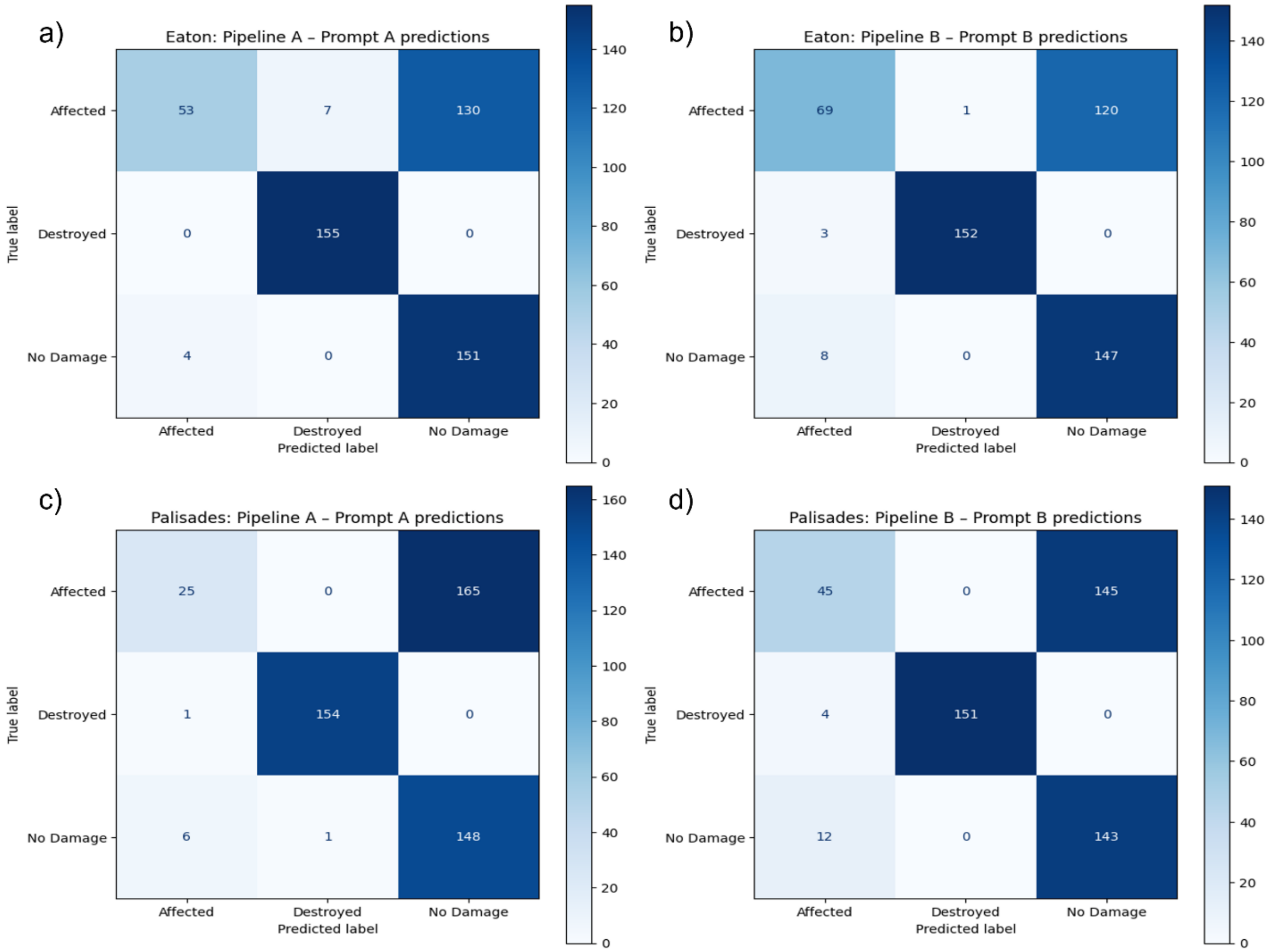

**Figure 5: Confusion matrix for the single-view test for Eaton with Pipeline A (5a), and Pipeline B (5b); Palisades with Pipeline A (5c) and Pipeline B (5d). The percentage of correctly classified households for *no damage*: 97.4% (5a), 94.8% (5b), 95.5% (5c), 92.3(5d). *Affected*: 27.9% (5a) 36.3% (5b) 13.2% (5c) 23.7% (5d). *Destroyed*: 100% (5a), 98.1% (5b), 99.4% (5c), 97.4% (5d).**

Table 6 shows the classification metrics of both pipelines for each study area. The F1 scores are used for comparison; however, precision, recall and Cohan Kappa are presented for transparency. The *affected* category had the lowest F1 scores for both pipelines in each study area, ranging from (0.225 to 0.511). This suggests the need for a workflow that can process multiple images for the same household as the damage could have occurred at another location of the household. These findings are similar to Luo and Lian (2024), as they found their *affected* category predictions from the trained ViT model to be the lowest of their three classification categories and the researchers acknowledge the need for a multi-image workflow (Luo & Lian, 2024).

**Table 6: Classification metrics for single front view images**

| Metrics | Eaton Pipeline A Prompt A | Eaton Pipeline B Prompt B | Palisades Pipeline A Prompt A | Palisades Pipeline B Prompt B |
|---|---|---|---|---|
| **Accuracy** | 0.718 | 0.736 | 0.654 | 0.678 |
| **Micro-Average F1 Score** | 0.718 | 0.736 | 0.654 | 0.678 |
| **Macro-Average F1 Score** | 0.700 | 0.732 | 0.617 | 0.663 |
| **Cohen's Kappa** | 0.587 | 0.611 | 0.495 | 0.528 |
| | | **No Damage** | | |
| **Precision** | 0.537 | 0.551 | 0.473 | 0.497 |
| **Recall** | 0.974 | 0.948 | 0.955 | 0.923 |
| **F1 Score** | 0.693 | 0.696 | 0.633 | 0.646 |
| | | **Affected** | | |
| **Precision** | 0.930 | 0.863 | 0.781 | 0.738 |
| **Recall** | 0.279 | 0.363 | 0.132 | 0.237 |
| **F1 Score** | 0.429 | 0.511 | 0.225 | 0.359 |
| | | **Destroyed** | | |
| **Precision** | 0.957 | 0.994 | 0.994 | 1.00 |
| **Recall** | 1.00 | 0.981 | 0.994 | 0.974 |
| **F1 Score** | 0.994 | 0.987 | 0.994 | 0.997 |

The main reason why this research decided to use a pre-trained MLLM model is to specifically classify the *affected* category better by processing multiple images for the same household. While a ViT can lead to higher F1 scores in some cases with a single image, training a model that can process multiple images is a rather cumbersome task. The input data would need to embed each image for the household and then average the embeddings. This approach is not fully practical for practitioners to utilize when a pre-trained model like a MLLM exists, as this is a time intensive task that may miss key details from the image. To examine the full capabilities of both Pipelines, section 3.3 examines the multi-view analysis.

### 3.3 Application of language models in a multi-view image analysis for damage assessments

The results from section 3.2. establish a baseline for the view comparison test and demonstrate the need to improve classification, specifically for the *affected* category. This research accomplishes this by using the same two pipelines and datasets but adding multiple views of houses when these views are present. Figure 6 shows the confusion matrix and highlights that the inclusion of different building perspectives resulted in a substantial increase in correctly classified buildings for the *affected* category, ranging from (77.4% to 95.5%.

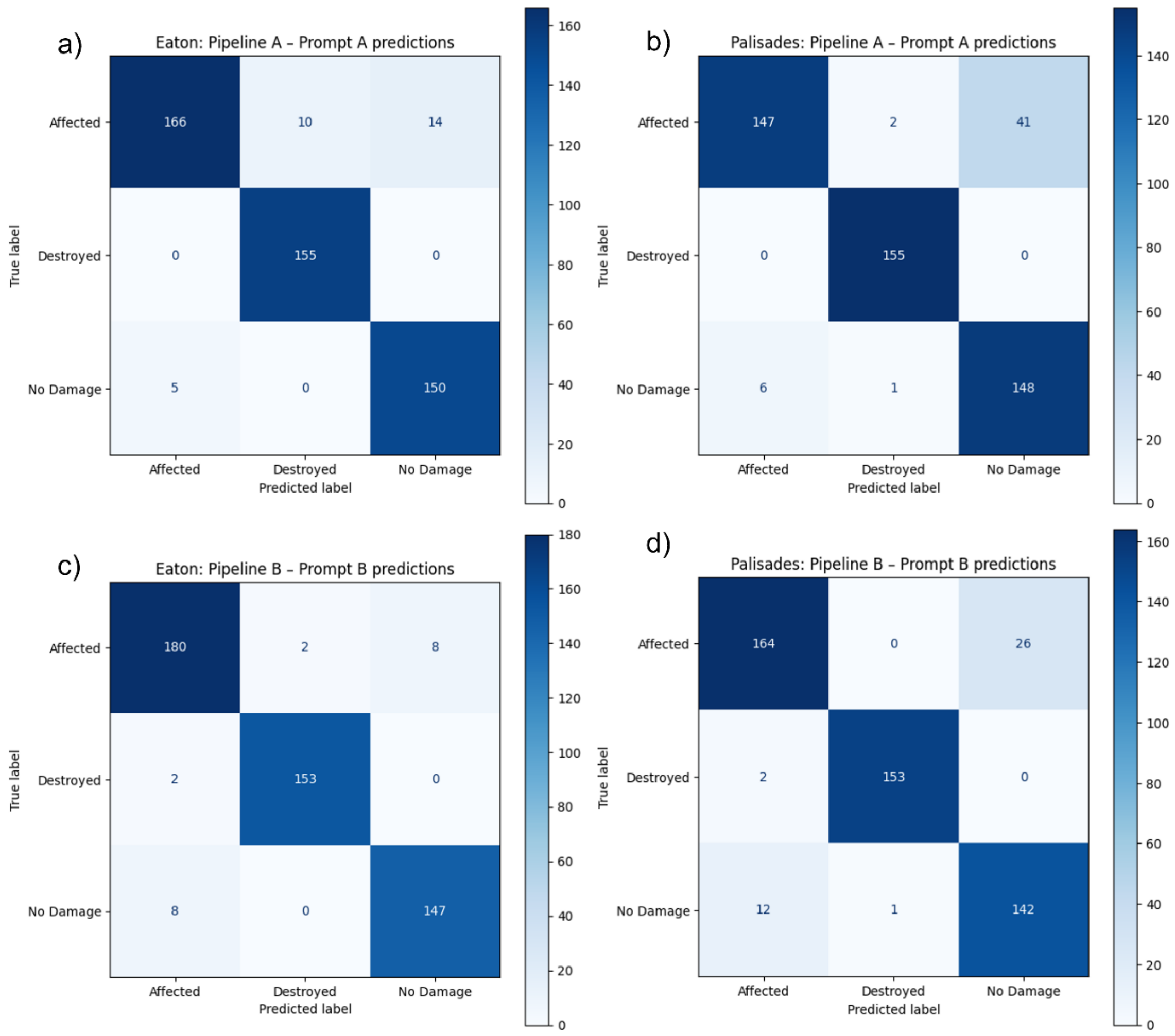

**Figure 6: Confusion matrix for the multi-view test for Eaton with Pipeline A (6a), and Pipeline B (6b); Palisades with Pipeline A (6c) and Pipeline B (6d). The percentage of correctly classified households for *no damage*: 96.8% (6a), 94.8% (6b), 95.5% (6c), 91.6(6d). *Affected*: 87.4% (4a) 94.7% (4b) 77.4% (6c) 95.5% (6d). *Destroyed*: 100% (6a), 98.1% (6b), 95.5% (6c), 98.7% (6d).**

Table 7 shows the classification metrics for both pipelines in each study area, when assessing multi-view images. The main finding of these results is that the *affected* category had an increase in their F1 scores when compared to table 6. This suggests that multi-view image assessment is critical for damage classification as the model can better capture damage visuals. Moreover, the *destroyed* category results were largely unaffected by adding multi-view images as most households for this category had only one image (see table 8). For *no damage*, this metric improved significantly as well, mostly because the majority of the misclassifications for affected, when examining the confusion matrix in figure 5, were allocated to the no damage category.

**Table 7: Classification metrics for multi-view images**

| Metrics | Eaton Pipeline A Prompt A | Eaton Pipeline B Prompt B | Palisades Pipeline A Prompt A | Palisades Pipeline B Prompt B |
|---|---|---|---|---|
| **Accuracy** | 0.942 | 0.960 | 0.900 | 0.960 |
| **Micro-Average F1 Score** | 0.942 | 0.96 | 0.900 | 0.960 |
| **Macro-Average F1 Score** | 0.943 | 0.961 | 0.903 | 0.961 |
| **Cohen's Kappa** | 0.913 | 0.940 | 0.850 | 0.877 |
| **No Damage Category** | | | | |
| **Precision** | 0.915 | 0.948 | 0.783 | 0.845 |
| **Recall** | 0.977 | 0.948 | 0.955 | 0.916 |
| **F1 Score** | 0.940 | 0.948 | 0.861 | 0.879 |
| **Affected Category** | | | | |
| **Precision** | 0.971 | 0.947 | 0.961 | 0.921 |
| **Recall** | 0.874 | 0.947 | 0.774 | 0.863 |
| **F1 Score** | 0.920 | 0.947 | 0.857 | 0.891 |
| **Destroyed Category** | | | | |
| **Precision** | 0.939 | 0.987 | 0.981 | 0.994 |
| **Recall** | 1.00 | 0.987 | 1.00 | 0.987 |
| **F1 Score** | 0.967 | 0.987 | 0.990 | 0.990 |

The results presented in this section confirm that the primary failure mechanism in section 3.2, for the *affected* category, was due to visual occlusion rather than the model's incapacity. Once both pipelines were provided with additional perspectives, they successfully identified damage cues that were previously hidden in the single-view experiment. Due to the improvement in the *affected* category, the *no damage* category also saw a metric boost, as the systematic false-positive error (where damaged homes were labeled as intact) was effectively eliminated. To further enhance these findings, section 3.4 presents the results of the McNemar's test and compares the class distribution as well as the ROC-AUC charts for both view cases.

### 3.4 An empirical comparison of the single-view and multi-view analysis

This section overviews the statistical results of the McNemar's test and compares the classification disparities of the single-view and multi-view analysis. The research preforms the McNemar's test on three cases. Case 1 is a comparison of Pipeline A's single view and multi-view assessment. Case 2 is similar to case 1 but with the results of Pipeline B. Case 3 compares the multi-view results of Pipeline A and B to compare a zero-shot prompting approach with a heuristic approach. Table 8 shows the McNemar's coefficient and p-values for the three cases and presents the number of households that have multiple images for each damage category to contextualize the statistical results.

**Table 8: The classification improvements MLLMs and LLMs provide in damage assessment**

| **Eaton** | | | **Palisades** | | |
|---|---|---|---|---|---|
| **Case 1: McNemar's test for Pipeline A's single-view and multi-view classification results** | | | | | |
| *No damage* | *Affected* | *Destroyed* | *No damage* | *Affected* | *Destroyed* |
| 111.08*** | 106.22*** | 1 | 118.20*** | 117.20*** | 1 |
| **Case 2: McNemar's test for Pipeline B's single-view and multi-view classification results** | | | | | |
| *No damage* | *Affected* | *Destroyed* | *No damage* | *Affected* | *Destroyed* |
| 106.22*** | 108.47*** | 1 | 108.64*** | 111.63*** | 1 |
| **Case 3: McNemar's test on the multi-view results for Pipeline A and Pipeline B** | | | | | |
| *No damage* | *Affected* | *Destroyed* | *No damage* | *Affected* | *Destroyed* |
| 0.266 | 3.56 | 1 | 0.124 | 2.06 | 1 |
| **Number of households with multiple images** | | | | | |
| *No damage* | *Affected* | *Destroyed* | *No damage* | *Affected* | *Destroyed* |
| 3 | 175 | 6 | 11 | 190 | 8 |

*p-value: *<0.01 **0.05 ***<0.0001*

The table shows the *no damage* and *affected* category's classification improvements are statistically significant for case 1 and case 2. However, for case 3, the improvements were not statistically significant. This shows that the baseline prompting methods in this study do not provide computational improvement; however, the exploration of other prompting methods is explored in section 3.6. For all three cases the *destroyed* category did not show any statistically significant improvement. There were 175 and 190 households with multiple images in Eaton and Palisades respectively. This is a contribution to the results of the McNemar's test. Most households that were *affected* were classified as *no damage* when using the single front façade images of the household. When the multi-view assessment was conducted, the MLLM workflow was capable of allocating most affected households to the correct category. While the *no damage* category had little households with multiple images, the improvement was statistically significant due to the correct classification of affected households. Section *S3* provides further context on the McNemar's results by highlighting the specific classification improvement across the single-view and multi-view case, while showing the little improvement when comparing Prompt A and Prompt B. Moreover, the findings are visually corroborated by the class distribution bar charts in figure 7, which display the single-view and multi-view results for both pipeline and prompt cases. In the single-view analysis, the predicted counts for the *affected* category are substantially lower than the ground truth of 190 buildings: Pipeline A – Prompt A identified only 57 (Eaton) and 32 (Palisades), while Pipeline B – Prompt B identified 80 (Eaton) and 61 (Palisades). Conversely, the *no damage* bars in the single-view charts are inflated, reflecting classifications that do not synergize with the ground truth label (GTL). However, the multi-view analysis reveals a distribution significantly closer to the GTL, confirming the statistical results in table 7. The bar charts for the *destroyed* category remain consistent across all four experiments, aligning with the lack of statistical variance observed in the McNemar test.

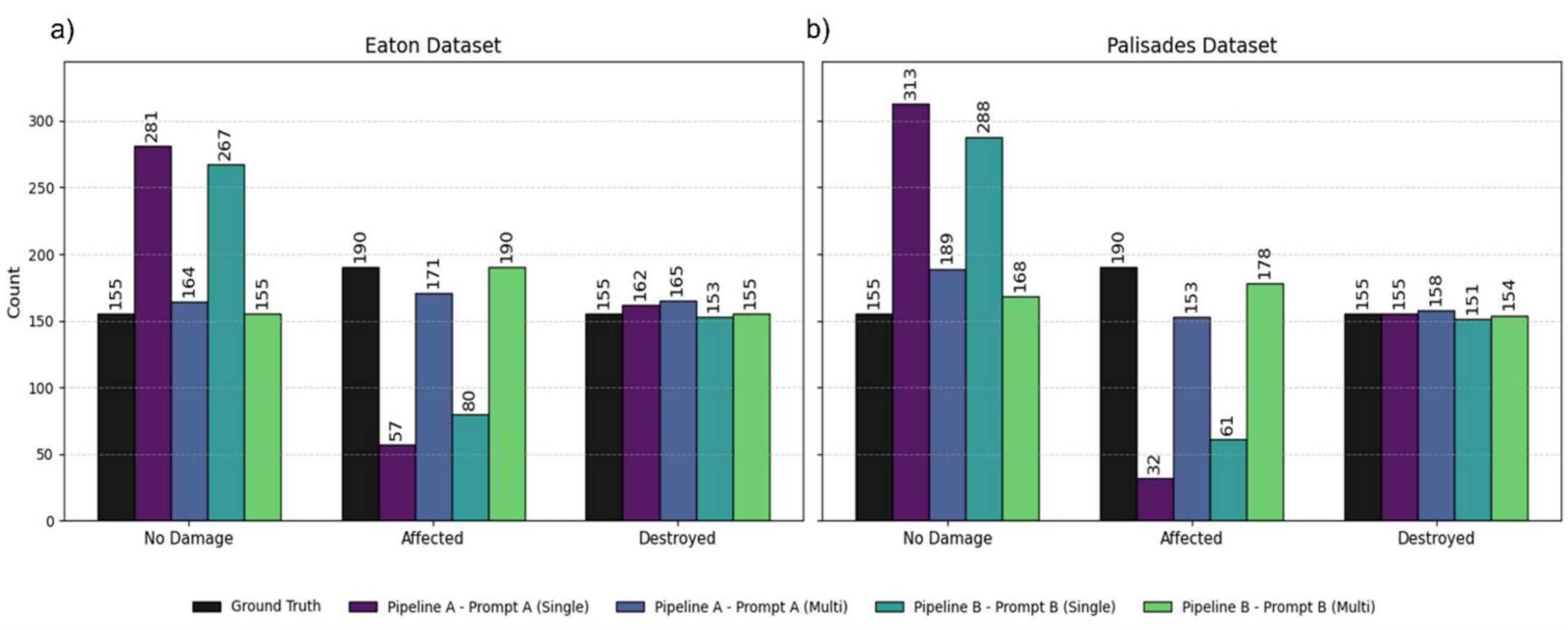

**Figure 7. Class distribution for single-view and multi-view assessments in (a) Eaton and (b) Palisades. Values represent counts for (Eaton; Palisades). Single-view: Pipeline A–Prompt A classified samples as *no damage* (281; 313), *affected* (57; 32), and *destroyed* (162; 155). Pipeline B – Prompt B showed increased sensitivity, resulting in *no damage* (267; 288), *affected* (80; 61), and *destroyed* (153; 151). Multi-View: Pipeline A – Prompt A yielded *no damage* (164; 189), *affected* (171; 153), and *destroyed* (165; 158). Pipeline B – Prompt B further refined detection to *no damage* (155; 168), *affected* (190; 178), and *destroyed* (155; 155).**

The ROC-AUC analysis (figure 8) provides a granular diagnostic of this performance shift. The single-view baselines demonstrate a collapse in discriminative capability for the *affected* category, with AUC values near the stochastic baseline of 0.5 (Pipeline A – Prompt A: 0.63 Eaton, 0.55 Palisades; Pipeline B – Prompt B: 0.66 Eaton, 0.59 Palisades). Geometrically, these single-view curves are linearized near the diagonal, reflecting the model's inability to distinguish damage when relying solely on frontal imagery. The integration of multi-view data restores the convexity of the curves, driving *affected* AUC scores into the 0.88–0.96 range. Crucially, the multi-view plots for both *affected* and *no damage* exhibit a sharp "elbow" characteristic approaching the ideal coordinate (0,1), confirming that the additional perspectives effectively minimize the trade-off between false positive and true positive rates.

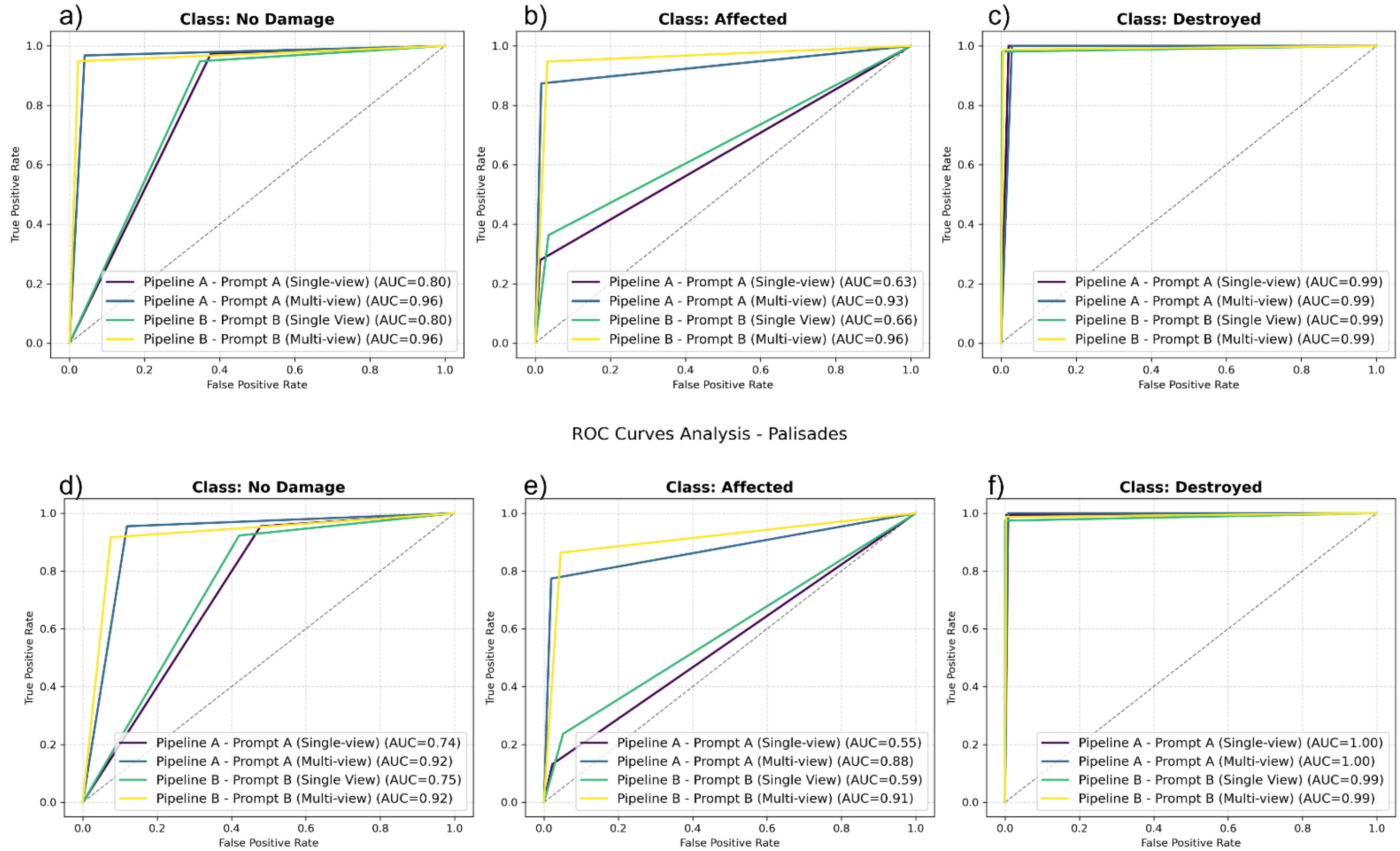

**Figure 8: ROC-AUC analysis in (a) Eaton and (b) Palisades. Values represent counts for (Eaton; Palisades). Single-view: Pipeline A – (Prompt A) classified samples as *no damage* (0.80; 0.74), *affected* (0.63; 0.55), and *destroyed* (0.99; 1). Pipeline B–Prompt B showed increased sensitivity, resulting in *no damage* (0.8; 0.75), *affected* (0.66; 0.59), and *destroyed* (0.99; 0.99). Multi-View: Pipeline A–Prompt A yielded *no damage* (0.96; 0.92), *affected* (0.93; 0.88), and *destroyed* (0.99; 1). Pipeline B– Prompt B) further refined detection to *no damage* (0.96; 0.92), *affected* (0.96; 0.91), and *destroyed* (0.99; 0.99).**

These statistical and distributional findings demonstrate the practicality of utilizing language models to enhance wildfire damage assessment. Specifically, the MLLM's capability to synthesize multiple images significantly improved classification in the *affected* category, addressing a critical gap in the existing literature. Moreover, because the model is pre-trained, it bypasses the need for complex data pre-processing—such as the averaging of image embeddings required by traditional architectures—to achieve similar feature extraction. This zero-shot workflow eliminates the need for training data, allowing practitioners to deploy the MLLM immediately after a natural hazard strikes. These results satisfy research objective one, as the multi-view MLLM workflow significantly enhanced classification performance for complex damage states.

To provide a granular understanding of why this improvement occurs, section 3.5 conducts a qualitative visual analysis of specific success and failure modes. This analysis focuses on the classification results from Pipeline B–Prompt B, as this configuration marginally outperformed Pipeline A–Prompt A across both view cases. Subsequently, to provide a more robust comparison of prompting strategies, section 3.6 compares the results among all four prompting cases within their respective pipelines. This comparison determines the extent to which prompting improves wildfire damage classification and interpretability. Specifically, this research examines how the inclusion of reasoning-based prompting approaches impacts classification results when compared to the baseline zero-shot (Prompt A) and heuristic (Prompt B) approaches. The experiments in section 3.6 are conducted exclusively on the multi-view image set, as the findings in this section have empirically established the superiority of holistic damage assessment.

### 3.5 A qualitative visual analysis of the single-view and multi-view classification results

Pipeline B, with its marginally better results, was the test case for the visual analysis, using randomly selected samples for both the single front view and multi-view. The first set of images were examples of Pipeline B's misclassification as shown in figure 9. The model classified these structures as *affected*. The structures, however, were clearly not livable; therefore, they were *destroyed*, according to the GTL. In all cases, the image has a standing structure which could be a cause for confusion for the model. Figure 10 (a–d) shows *destroyed* structures that were correctly classified in both single-view and multi-view. In contrast, in figure 9, the destroyed structures were standing to some degree, thus a cause for confusion in the model. This provides an opportunity to finetune LLM prompts to capture cases in which unlivable homes are still standing from the fire.

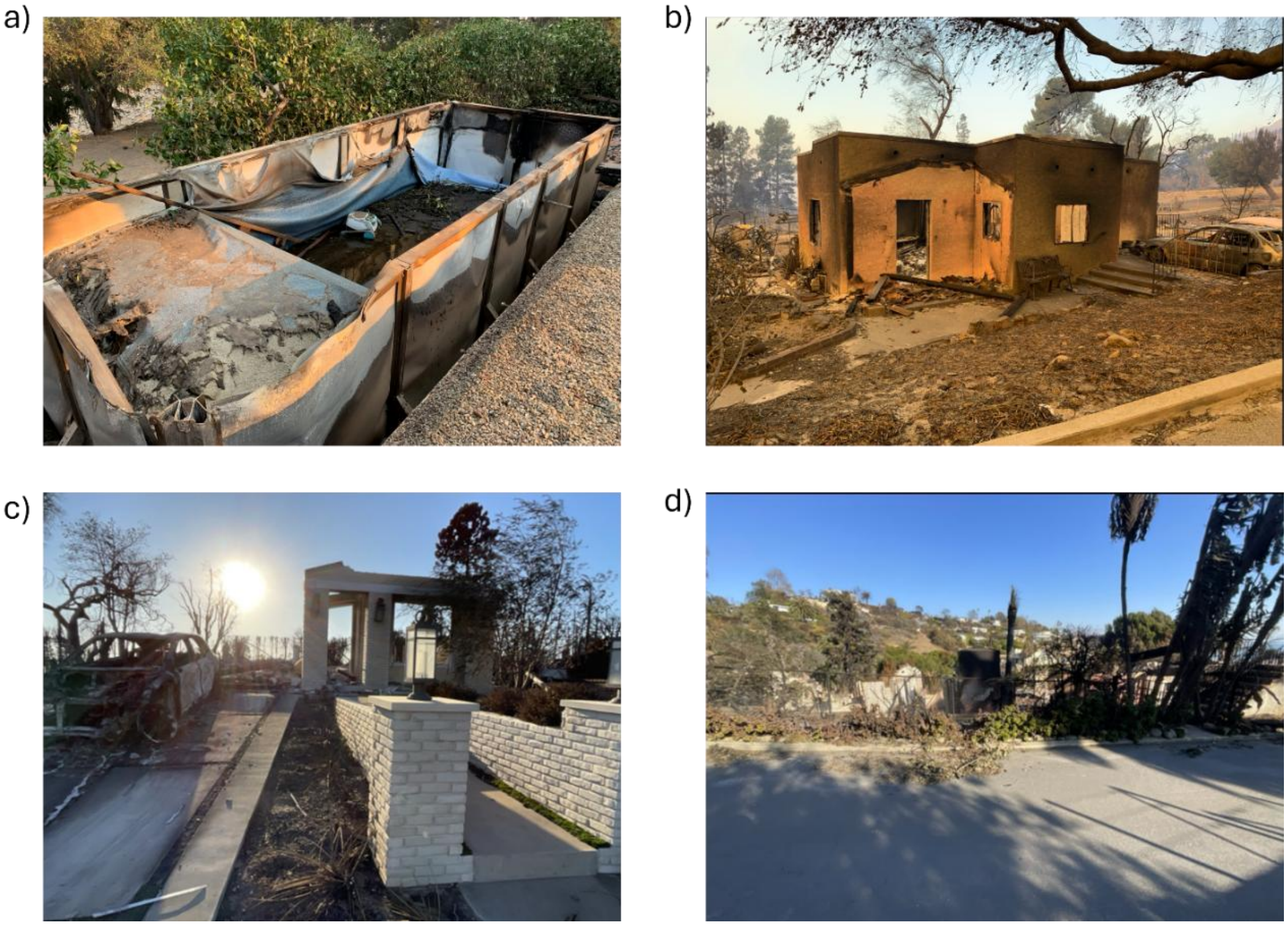

**Figure 9: All homes in this figure only had a single-view. Pipeline B's misclassification examples for *destroyed (GTL)* buildings: 9a (Palisades), 9b (Palisades), 9c (Eaton) and 9d (Eaton) were all classified as *affected* due certain components of the building standing.**

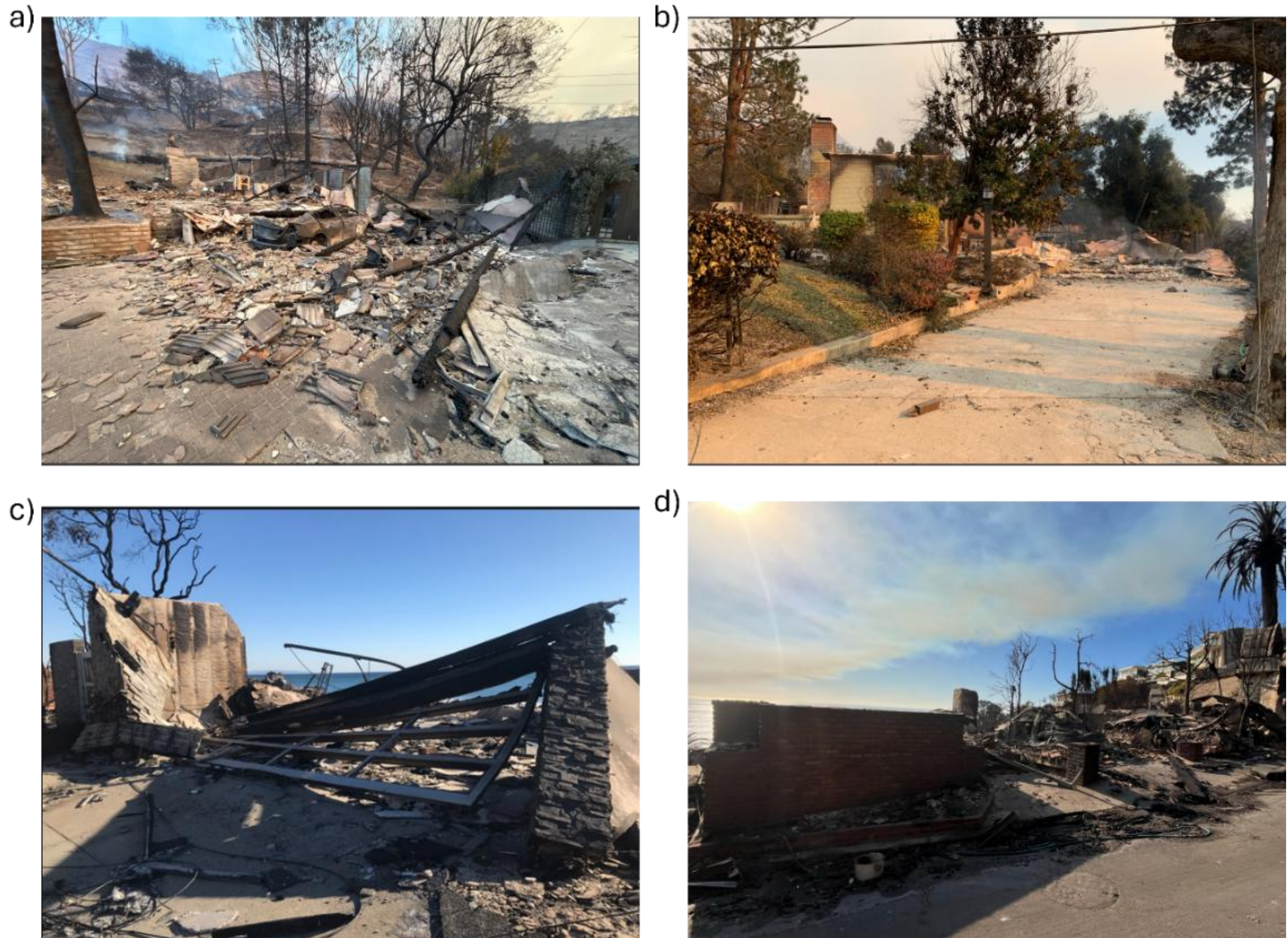

**Figure 10: All homes in this figure only had one single-view. Pipeline B correctly classified figure 10 (a–d) as *destroyed* which is the same as the GTL. Figure 10a and 10b are in Eaton. Figure 10c and 10d are in Palisades.**

Figure 11 shows cases where the single-view only images (a), (c), (e), and (g) were classified as *no damage*; however, when the multi-view assessment was conducted, the household was classified as *affected*. Figure 11 (b) (d), (f), and (h) are the alternative views of houses (a), (c), (e), and (g), respectively. This shows the importance of having a workflow that is capable of processing multiple images.

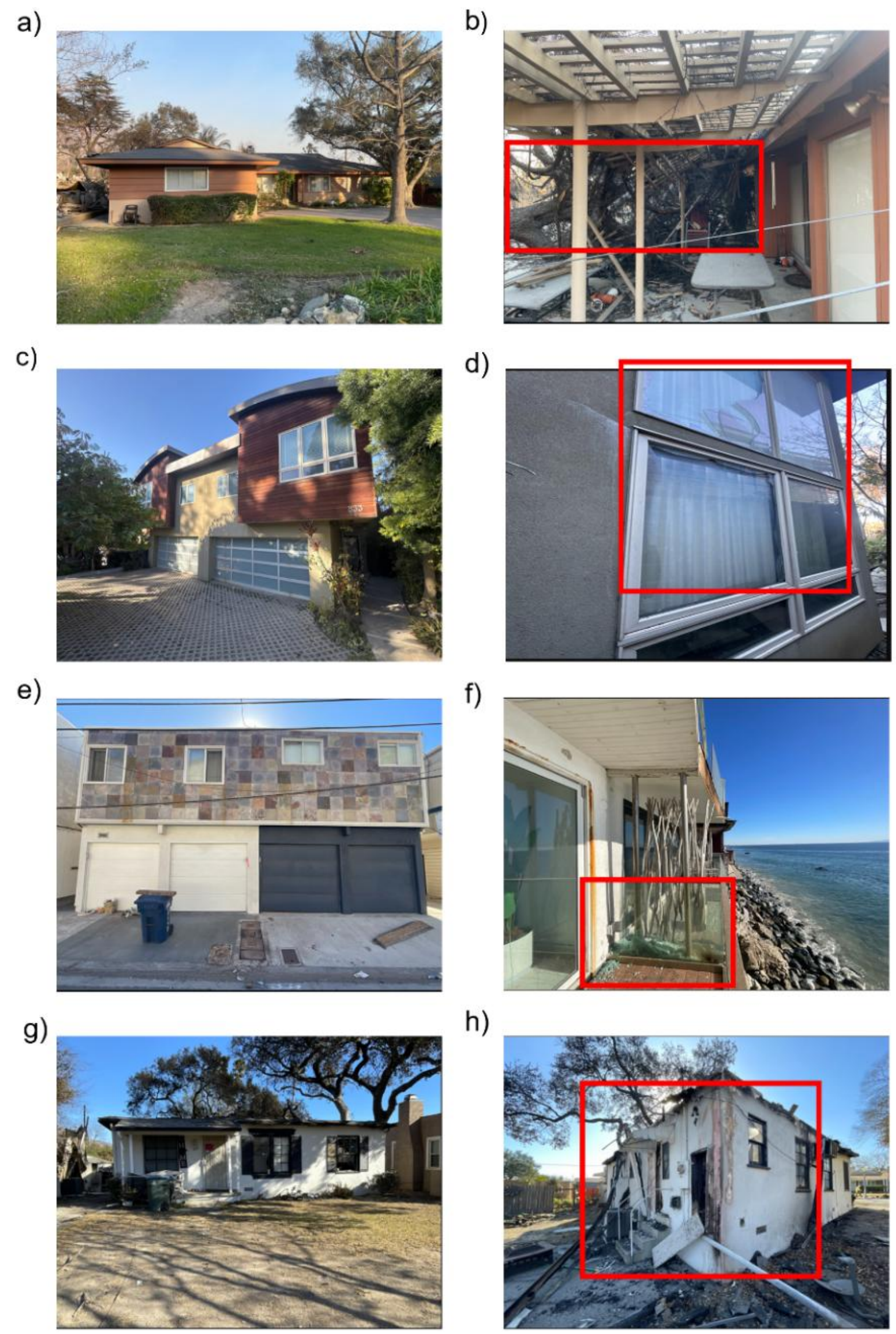

**Figure 11: In each row are pairs of images for the same house; front views of houses ((a), (c), (e), (g)) were classified as *no damage*. In the multi-view assessment, (a)–(b); (c)-(d); (e)-(f); and (g), (h) were inputs for the Pipeline B. They were classified as *affected* which matches the GTL.**

Figure 12 shows cases where the household was misclassified for both pipelines; (a) and (b) are from the same household. Both models predicted *no damage*; however, broken glass is evident in figure 12 (b) that caused ground truth to be classified as *affected*. The same can be seen in figure 12 (d), (e), and (f). These images from the same house, which were labeled as *no damage* for both views. The single-view does not show any damage; however, broken glass is visible in Figure 12 (e) and (f), which was difficult for the model to identify, despite the prompt regarding the presence of broken glass. Future research can perhaps alter the brightness of the image to make the broken glass more identifiable. Figure 12 (c) had a GTL of *no damage*, but the object in front of the house led both views to predict damage since the prompt of regarding debris in front of the house was true. This demonstrates a tradeoff, as some images could have objects, such as tarp in this case, in front of the house and the model may classify the household as *affected*.

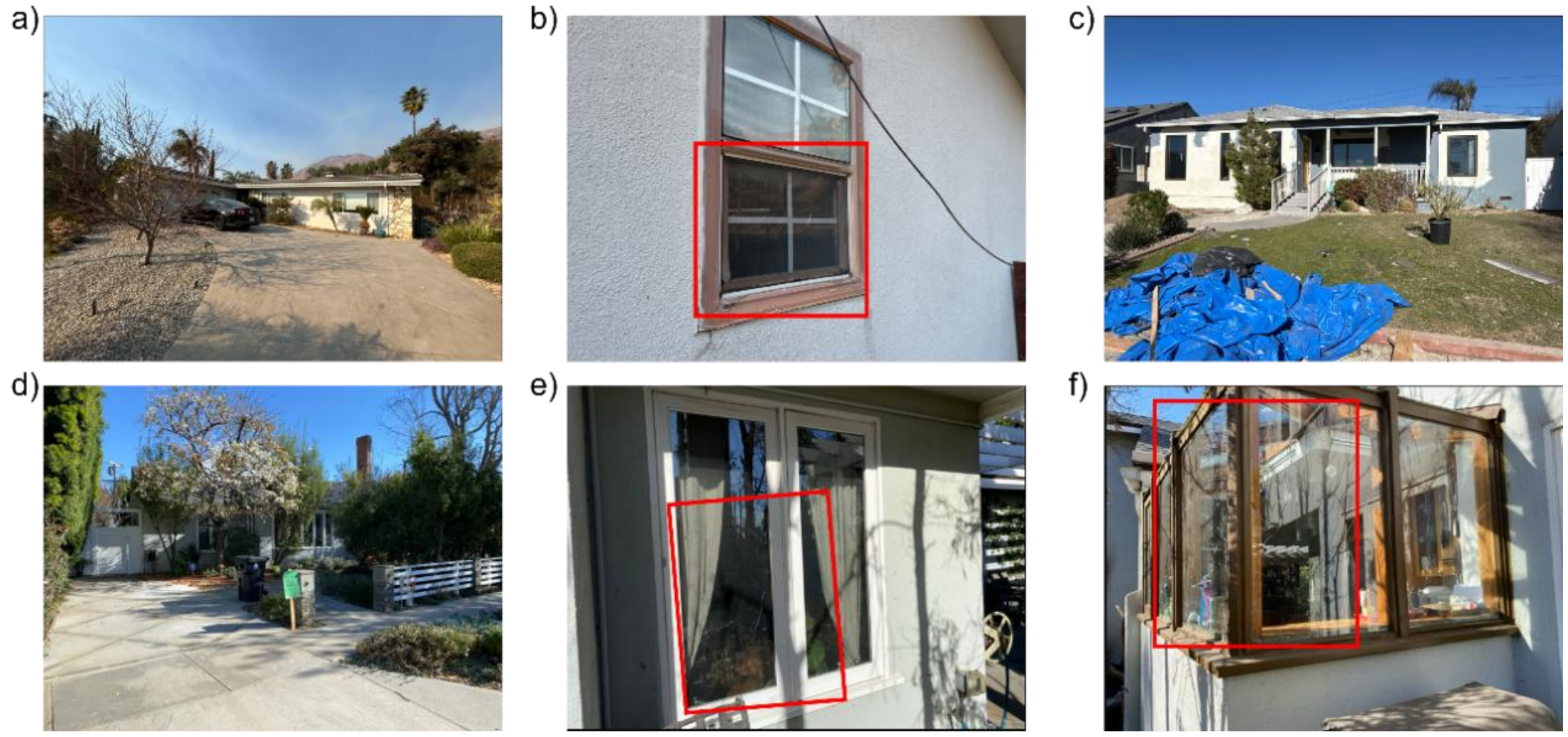

**Figure 12: The household in 12 (a – b) and 12 (d – f) were classified as no damage, but the ground truth was affected due to the broken glass. The household in 12 (c) was classified as affected due to the tarp being interpreted by debris; however, the GTL was no damage.**

This visual analysis complements the findings in section 3.2 and section 3.3, specifically figure 11 shows the value of a multi-view image assessment. There are some cases where the model was not able to properly identify damage, specifically with broken glass as seen in figure 12. This limitation can be addressed in future studies with prompt engineering. Overall, the visual analysis was able to highlight how language models can improve damage classification.

### 3.6 Exploring the application of different prompt methods for the multi-view analysis

Building on the findings from section 3.4, which established the superiority of the multi-view approach, this section addresses the second research objective by exploring the impact of reasoning based prompting strategies. Prompt C and Prompt D were tested using the multi-view configuration due to its classification improvement for the *affected* and *no damage* categories.

Table 9 and table 10 present the classification metrics for Eaton and Palisades, respectively, the different prompting experiments. In general, the reasoning-based prompting methods yielded slightly lower classification metrics compared to the baseline prompts, Prompt A and Prompt B, across both study areas. However, a notable exception was observed in the Palisades Pipeline A – Prompt D test, where the *affected* category achieved an F1 score of 0.900, while the *no damage's* F1 score was 0.928. Both of these scores were the highest scores recorded among all Palisades configurations. Conversely, the Pipeline B – Prompt D configuration demonstrated a decline in performance. The F1 score for the *affected* category dropped to 0.828 in Eaton and 0.808 in Palisades, the lowest values recorded out of the four additional tests. This degradation occurs because the text-based model is required to classify damage based on the vision model's unstructured reasoning (Prompt D) rather than the rigid, structured indicators provided in the heuristic approach (Pipeline B-Prompt B). This comparison highlights that while decoupled text models excel at processing structured data, they struggle to interpret verbose, unstructured reasoning without introducing classification errors. When comparing the *affected* category F1 scores amongst Prompts A, C and D within Pipeline A, they are generally in the same range for both Eaton (Prompt A: 0.920; Prompt C: 0.908; Prompt D: 0.900) and Palisades (Prompt A: 0.857; Prompt C: 0.845; Prompt D: 0.900). Given this metric stability, the selection of a prompting strategy is defined not by accuracy, but by the operational constraints of the deployment environment (table 5). The zero-shot baseline (Prompt A) offers a high-throughput, low-latency solution suitable for rapid initial triage, yet it operates as a "black box." Conversely, Prompt C and Prompt D incur a higher computational cost (table 4), but provide auditable reasoning chains. This interpretability transforms the model from a simple classifier into a transparent decision-support tool. This allows practitioners to gain interpretability that could enhance their decision making and planning after the wildfire occurs.

**Table 9: Classification metrics for multi-view images for all pipeline prompt pairs in Eaton**

| Metrics | Eaton Pipeline A-Prompt A | Eaton Pipeline B-Prompt B | Eaton Pipeline A-Prompt C | Eaton Pipeline A-Prompt D | Eaton Pipeline B-Prompt D |
|---|---|---|---|---|---|
| **Accuracy** | 0.942 | 0.960 | 0.932 | 0.920 | 0.836 |
| **Micro-Average F1 Score** | 0.942 | 0.96 | 0.932 | 0.920 | 0.836 |
| **Macro-Average F1 Score** | 0.943 | 0.961 | 0.933 | 0.923 | 0.821 |
| **Cohen's Kappa** | 0.913 | 0.940 | 0.897 | 0.878 | 0.830 |
| **No Damage Category** | | | | | |
| **Precision** | 0.915 | 0.948 | 0.886 | 0.940 | 0.940 |
| **Recall** | 0.977 | 0.948 | 0.955 | 0.916 | 0.903 |
| **F1 Score** | 0.940 | 0.948 | 0.919 | 0.928 | 0.921 |
| **Affected Category** | | | | | |
| **Precision** | 0.971 | 0.947 | 0.933 | 0.853 | 0.730 |
| **Recall** | 0.874 | 0.947 | 0.884 | 0.952 | 0.903 |
| **F1 Score** | 0.920 | 0.947 | 0.908 | 0.900 | 0.828 |
| **Destroyed Category** | | | | | |
| **Precision** | 0.939 | 0.987 | 0.980 | 1 | 1 |
| **Recall** | 1.00 | 0.987 | 0.954 | 0.889 | 0.828 |
| **F1 Score** | 0.967 | 0.987 | 0.919 | 0.934 | 0.714 |

**Table 10: Classification metrics for multi-view images for all pipeline prompt pairs in Palisades**

| Metrics | Palisades Pipeline A-Prompt A | Palisades Pipeline B-Prompt B | Palisades Pipeline A-Prompt C | Palisades Pipeline A-Prompt D | Palisades Pipeline B-Prompt D |
|---|---|---|---|---|---|
| **Accuracy** | 0.900 | 0.960 | 0.888 | 0.920 | 0.839 |
| **Micro-Average F1 Score** | 0.900 | 0.960 | 0.888 | 0.920 | 0.839 |
| **Macro-Average F1 Score** | 0.903 | 0.961 | 0.892 | 0.922 | 0.838 |
| **Cohen's Kappa** | 0.850 | 0.877 | 0.888 | 0.92 | 0.753 |
| | | | **No Damage Category** | | |
| **Precision** | 0.783 | 0.845 | 0.760 | 0.940 | 0.820 |
| **Recall** | 0.955 | 0.916 | *0.961* | 0.916 | 0.968 |
| **F1 Score** | 0.861 | 0.879 | 0.845 | 0.928 | 0.888 |
| | | | **Affected Category** | | |
| **Precision** | 0.961 | 0.921 | *0.942* | 0.854 | 0.784 |
| **Recall** | 0.774 | 0.863 | 0.767 | 0.953 | 0.834 |
| **F1 Score** | 0.857 | 0.891 | 0.845 | 0.900 | 0.808 |
| | | | **Destroyed Category** | | |
| **Precision** | 0.981 | 0.994 | 1 | 1 | 1 |
| **Recall** | 1.00 | 0.987 | 0.961 | 0.884 | 0.689 |
| **F1 Score** | 0.990 | 0.990 | 0.980 | 0.938 | 0.817 |

The class distribution results (figure 13) for all five prompting tests provide further context into the classification metrics. Specifically, the Pipeline B – Prompt D test shows how the *affected* category is overclassified (254 vs 190 in Eaton and 204 vs 190 in Palisades). Consequently, this test underestimates the number of destroyed buildings (97 vs 155 in Eaton and 112 vs 155 in Palisades). This can be problematic in post-damage assessment as the classification of destroyed buildings is paramount for resource allocation. Therefore, when utilizing this workflow, it is critical that both visual cues and prompting are used concurrently. Pipeline A with Prompt A, C, and D generally have little skewness among all three damage classes. However, a decoupled architecture can offer a token-efficient approach in cases where only the classification criteria for damage needs to be updated, and the visual descriptions generated in the initial pass are stored.

Class label distribution: prompt testing

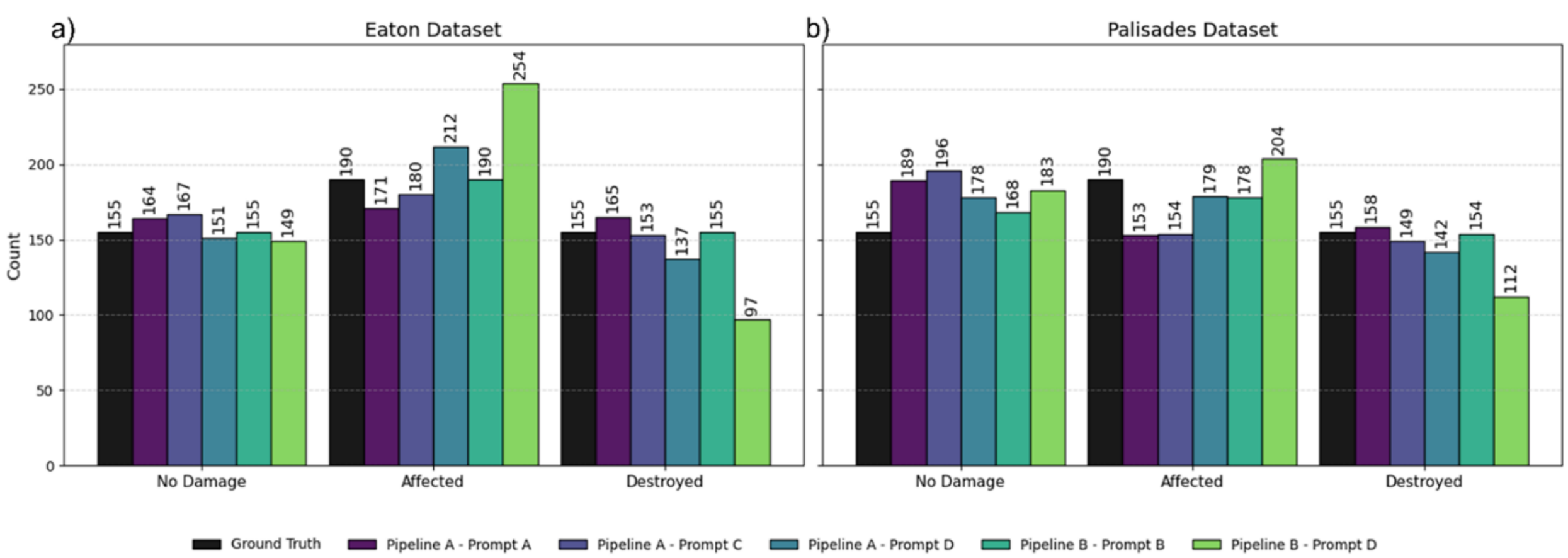

**Figure 13: Class distribution for multi-view assessments in (a) Eaton and (b) Palisades. Values represent counts for (Eaton; Palisades). Pipeline A – Prompt A, C, D classified samples as *no damage* (164, 151, 167; 189, 178, 183), *affected* (164, 161, 167; 153, 179, 204), and *destroyed* (165, 137, 153; 155, 142, 149). Pipeline B – Prompt B, D classified *no damage* (155, 149; 168, 189), *affected* (190, 254; 178, 204), *destroyed* (155, 97; 154, 142).**

The ROC-AUC charts in figure 14 provide a geometric assessment of the potential loss in discriminative fidelity incurred by utilizing reasoning prompts. For all classification categories in both study areas for Pipeline A, the ROC trajectories for Prompt A, C, and D exhibit topological convergence. All three strategies maintain high convexity with sharp elbows approaching the ideal coordinate (0,1). However, when examining Pipeline B – Prompt D there is a discriminative collapse with the *destroyed* category as the curve is closer to the stochastic baseline (0.5) as a result of the overestimation of the *affected* case. This characteristic has not been observed in either prompting strategy and even in the single-view analysis in figure 8, the destroyed category was nowhere near the baseline of 0.5. This linearization quantifies the "hallucination gap," visualizing exactly where the text model loses precision when forced to interpret unstructured visual descriptors. These results show a limitation in decoupled architectures, such as Pipeline B. While Pipeline B performed well with rigid Heuristics (Prompt B), it failed when attempting complex S-CoT reasoning (Prompt D). This indicates that when a text model is separated from the visual input, it loses the nuance required to validate reasoning; it effectively "hallucinates" based on the text description rather than verifying the image.

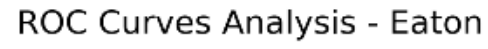

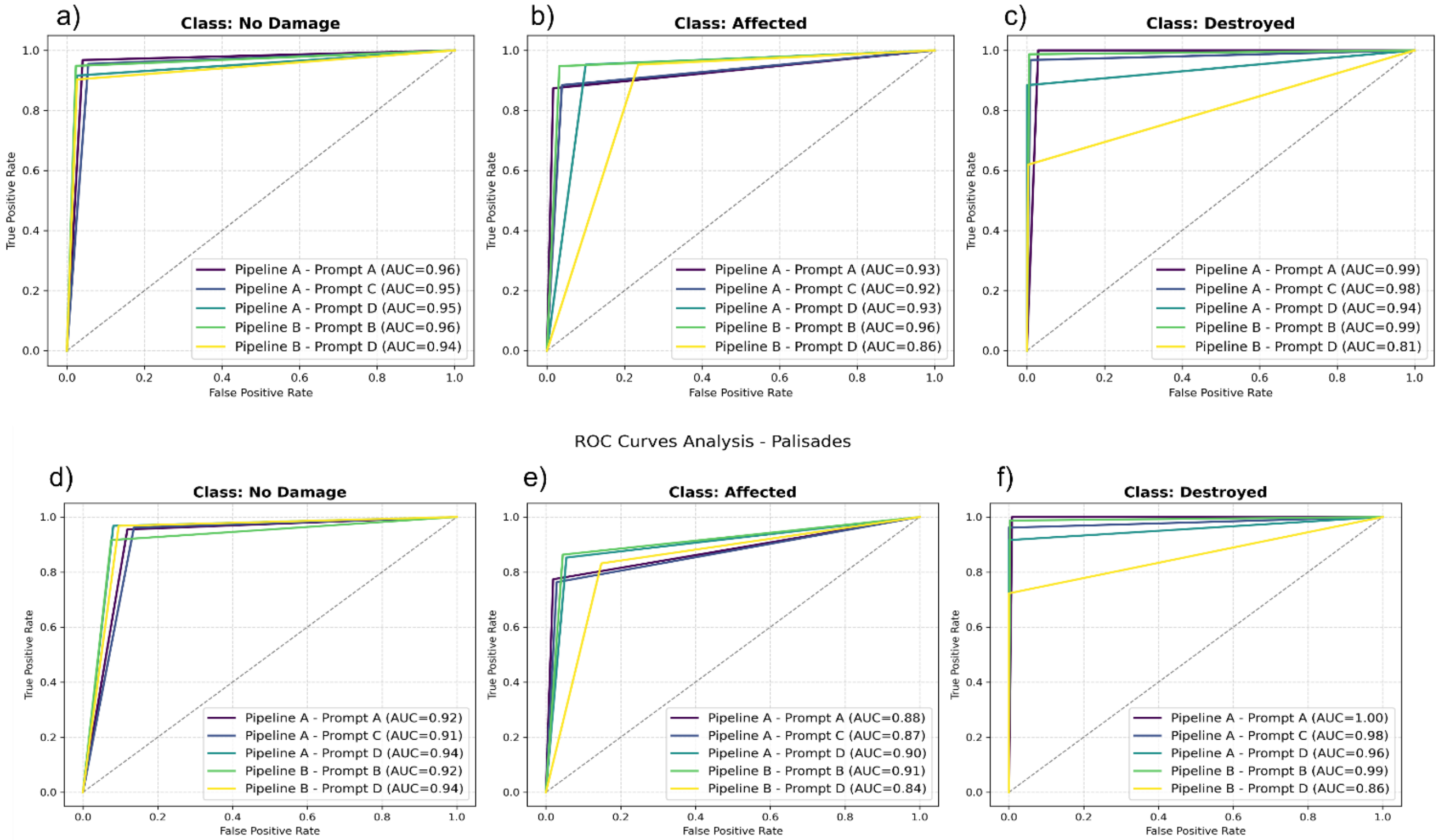

**Figure 14. ROC-AUC analysis multi-view assessments in Eaton (a–c) and Palisades (d–f). Values represent counts for (Eaton; Palisades). Pipeline A–Prompt A, C, D classified samples as *no damage* (0.92, 0.94, 0.91; 0.96, 0.95, 0.95), *affected* (0.88, 0.87, 0.90; 0.93, 0.92, 0.93), and *destroyed* (1, 0.98, 0.96; 0.99, 0.98, 0.94). Pipeline B–Prompt B, D classified *no damage* (0.92, 0.94; 0.96, 0.94), *affected* (0.91, 0.84; 0.96, 0.86), *destroyed* (0.99, 0.86; 0.99, 0.81).**

The results presented in this section demonstrate that while reasoning prompting strategies do not outperform the baseline prompts in classification accuracy, they have a profound impact on model stability and interpretability. For the integrated Pipeline A, the results indicate that the choice between Prompt A, C, and D is not a trade-off in accuracy, but in operational utility. As detailed in the computational resource analysis (table 4 and table 5), reasoning-based prompts incur significantly higher latency and token costs. While negligible for a sample of 500 households, this cost disparity becomes a critical bottleneck when scaling to full datasets of ~12,000 or ~18,000 parcels. Moreover, a clear hallucination phenomenon emerges when a text-based classifier is forced to interpret unstructured visual descriptors. In the case of Eaton and Palisades, the Pipeline B–Prompt D test overclassified the *affected* categories at the expense of correctly identified destroyed buildings, which can be a significant bottleneck in disaster recovery. However, this bottleneck can be mitigated if the use of a text based classifier is required, as a list of rigid damage indicators (Pipeline B–Prompt B) better identifies affected and destroyed buildings. Consequently, practitioners must weigh the immediate use case depending on if the situation demands a rapid, zero-shot assessment to prioritize speed and life–saving triage, or if the situation requires auditable interpretability to support long-term recovery planning.

In conclusion, research objective two is accomplished by demonstrating that prompting strategies fundamentally impact damage assessment through a trade–off between operational utility and interpretability rather than enhancing classification accuracy. The results in table 9 and 10 show no noticeable increase in the classification metrices across the prompting methods and table 8's McNemar results show no statistical advantage between Prompt A and Prompt B. However, when exploring the prompting methods beyond classification accuracy, there are advantages and disadvantages. Within Pipeline A, and all prompt pairs there exists performance parity, allowing practitioners to select a strategy based on resource availability and the need for transparency without compromising accuracy. Conversely, the results highlight a critical limitation in decoupled architectures: when Prompt D is implemented in Pipeline B, accuracy significantly degrades. This failure stems from the text model's inability to process abstract reasoning without supporting visual cues. Therefore, when text-based classification is required, a rigid, indicator–based approach (Prompt B) is strictly superior. To further examine these dynamics, section 3.7 presents a visual analysis of the model's reasoning outputs, offering qualitative evidence to complement these quantitative findings.

### 3.7 A qualitative visual analysis comparison of prompting methods based on the multi-view perspective

To complement the findings in section 3.6, a qualitative, multi-view only, visual analysis is presented to demonstrate the reasoning logic of Prompt C and Prompt D. The GTL for figures 15–17 is *affected.* Figure 18 presents four cases where Pipeline B–Prompt D misclassifies the ground truth *destroyed* buildings as *affected*. The model's reasoning is summarized for each figure in this section, but the reasoning outputs can be seen in the supplementary material section (*S5*).

Figure 15 shows cases where all pipeline and prompt pairs correctly the parcels as *affected*. In these cases, the images on the right-hand side (b,d,f,h) show heat damage, such as soot staining, blister paint, or charred beams/eaves. The model's justifications for correctly categorizing these buildings as affected generally was due to no structural damage or breaching of the building's envelope; however, there was clear presence of cosmetic or structural heat damage.

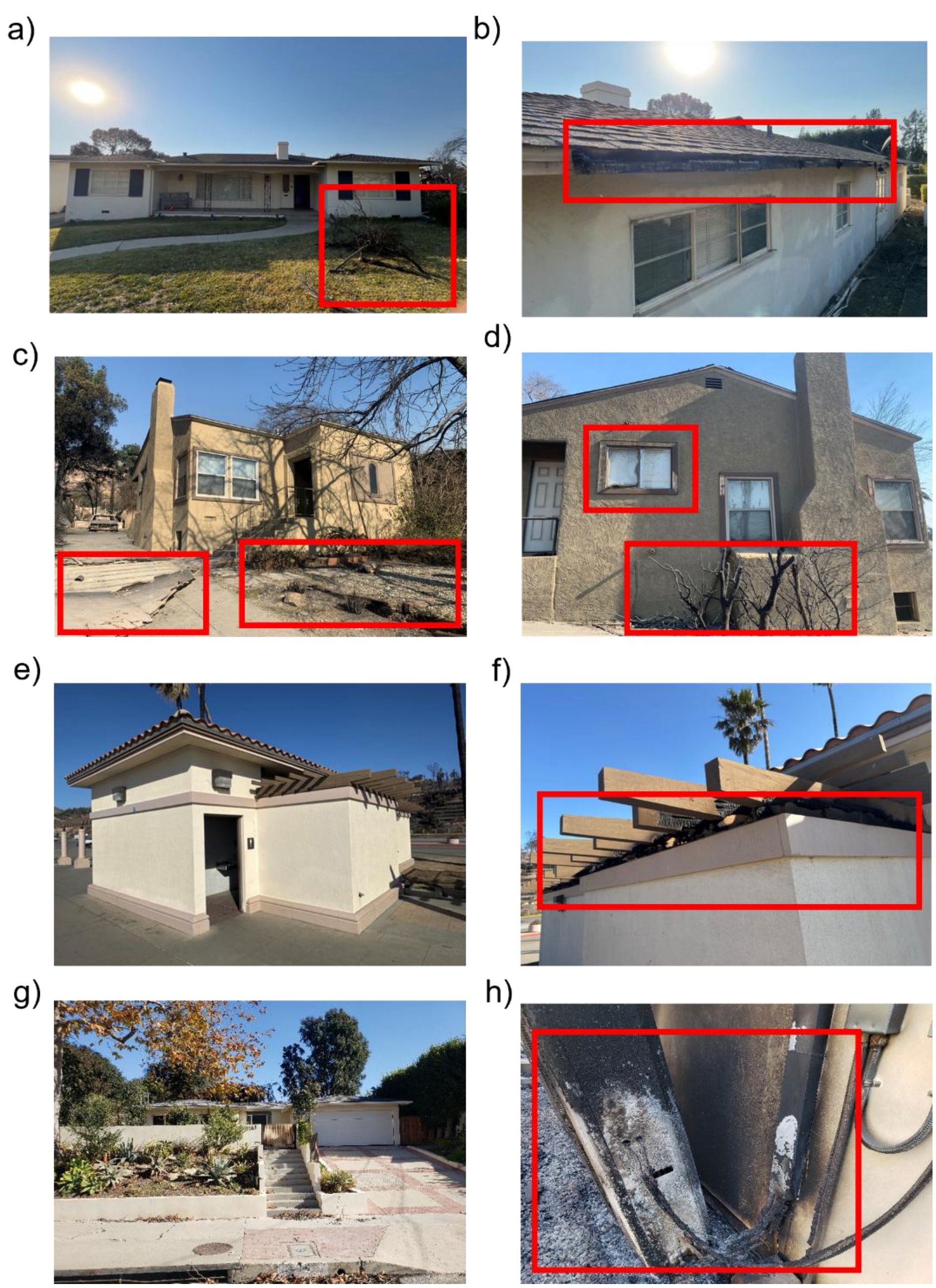

**Figure 15: Each individual row is a pair of images for the same building in Eaton (a – d) and Palisades (e – h). All pipeline prompt pairs correctly classified these buildings as *affected* which is the same as the GTL.**

Figure 16 shows a mixture of classifications. Figure 16a –16b and figure 16c – 16d both had Pipeline B – Prompt B misclassify these buildings in Eaton as *no damage.* Prompt B could not detect the damage to the fence (figure 16b) or deck (figure 16d), due to the rigid nature of the checklist. The same figures were misclassified as *no damage* in Pipeline A – Prompt C because the prompt mostly focused on the main building envelope or cosmetic damage. The other three pipeline prompt pairs were able to identify these cases due to the zero-shot's ability to simplify the problem and the S-CoT's multiple thinking paths. Figure 16e – 16f only had Pipeline B – Prompt B misclassify these buildings in Palisades due to the checklist not accounting for the subtle cosmetic damage. This highlights that, while a check-list approach is valuable, there needs to be a balance of identifying true cosmetic damage and noise in the picture. The zero-shot prompt identified this damage due to its simple approach. Both reasoning methods across all pipelines identified these damages and reasoning prompts correctly classified these buildings due to SC's iterative approach and S-CoT's multiple thinking paths.

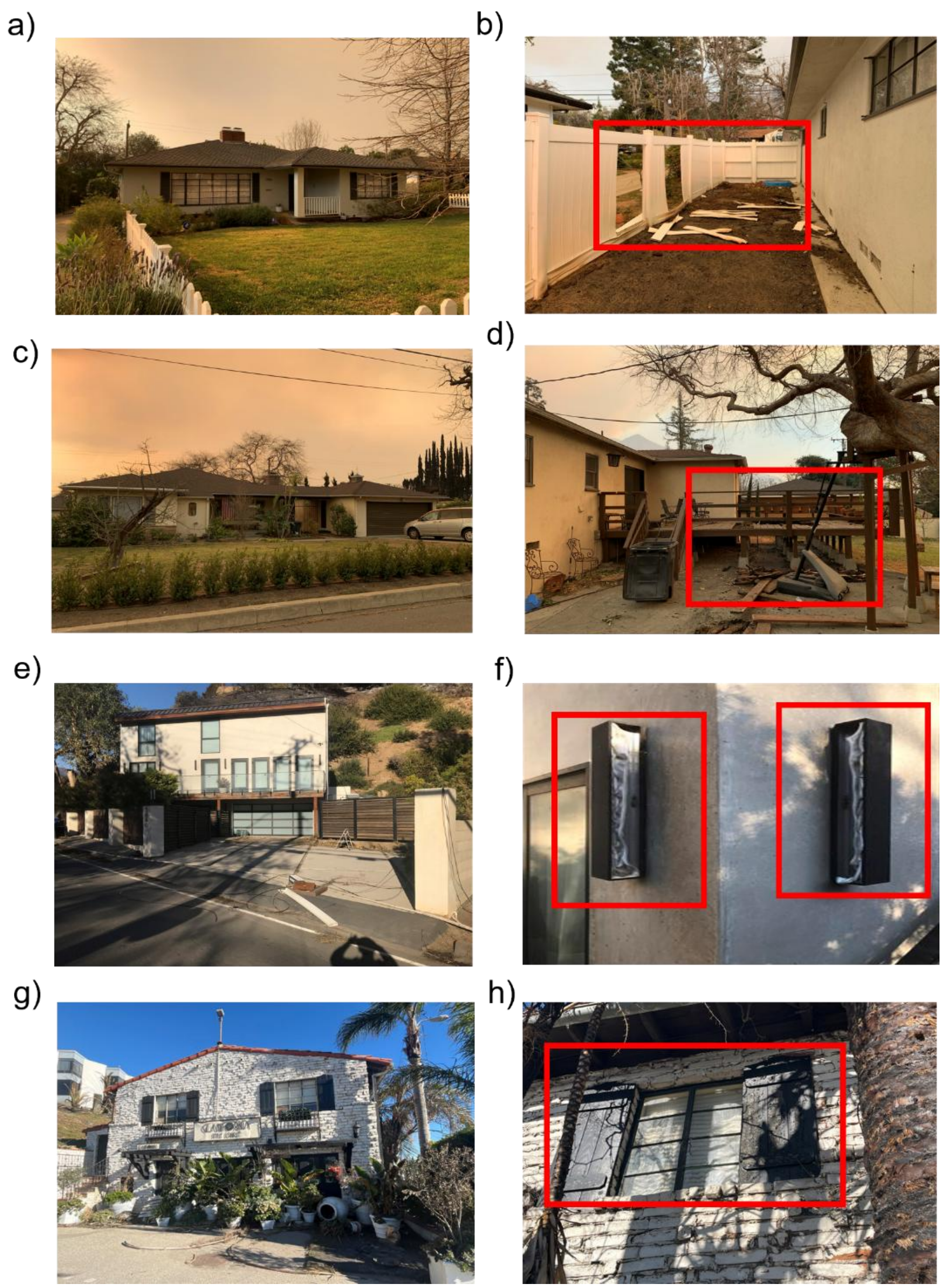

**Figure 16: Each individual row is a pair of images for the same building in Eaton (a – d) and Palisades (e – h). This is a mixed result classification where some prompts successfully classified images and others did not. For example, Pipeline A – Prompt C and Pipeline B – Prompt B misclassified 16a – 16b and 16c–16d as *no damage*, while the other prompts correctly classified the buildings. In 16e – 16f and 16g – 16h, only Prompt B misclassified the images, while the remaining were correct. The GTL for these buildings were *affected.***

Figure 17 shows four cases where all prompting strategies failed to correctly classify the structures. These misclassifications primarily stem from the inherent difficulty vision models face when identifying broken glass, particularly when the glass is not the primary focal point of the image. As seen in figures 17b and 17d, the model successfully identifies unscathed structural components but fails to detect the broken glass that is clearly evident to a human inspector. Mathematically, this represents a significant challenge in pixel-value differentiation. Due to the transparent nature of glass, the pixel values captured by the sensor, which represent the interior of the building, often remain similar whether the glass is intact or missing. Consequently, the model cannot distinguish between a clear windowpane and a structural breach. Furthermore, figures 17f and 17h demonstrate how physical obstructions, such as roof overhangs (17f) or dense vegetation(17h), can further obscure critical damage markers. This analysis underscores the importance of input data quality; had the focal point of these images been oriented toward the structural damage, the model's chances of achieving a correct classification would be better.

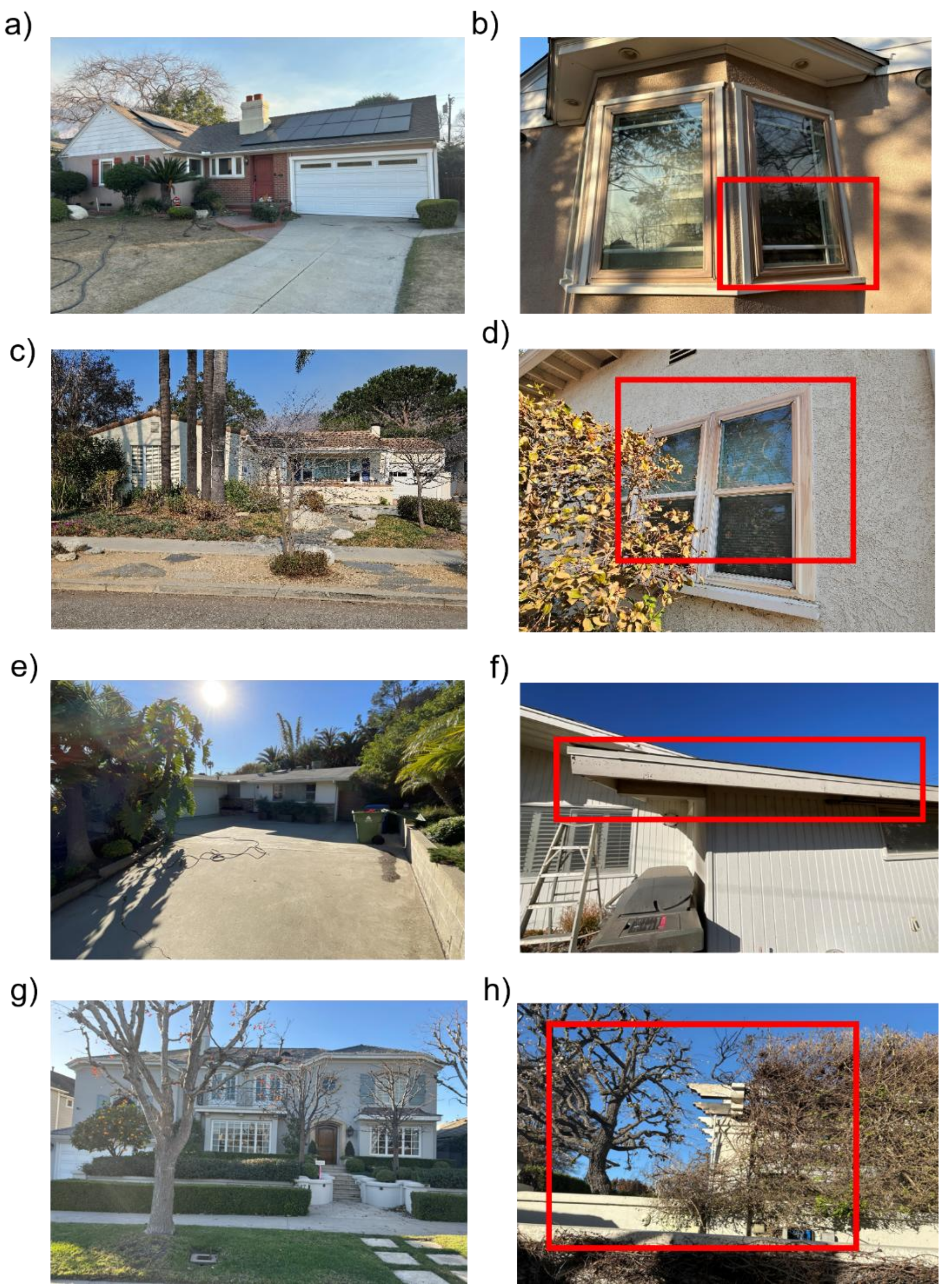

**Figure 17: Each individual row is a pair of images for the same building in Eaton (a – d) and Palisades (e – h). All pipeline prompt pairs misclassified these buildings as *no damage* while the GTL was *affected*.**

Figure 18a – d show four cases where Pipeline B – Prompt D misclassified all buildings as *affected* rather than *destroyed.* In the supplementary material section, the model was able to identify the extensive fire impact, such as debris, melted materials, missing walls, missing roofs, and a complete collapse. However, with no visual indicators to support the claim the model underestimated the severity of damage. To a human, if they were presented with the same textual information, the key indicators of a missing roof and wall would lead the individual to classify the building as destroyed since it is not livable. This highlights two things, one, when using text only classification a rigorous definition of the damage state must be defined. Second, and more importantly, it is critical to have a human-in-the-loop system that can capture these critical mistakes.

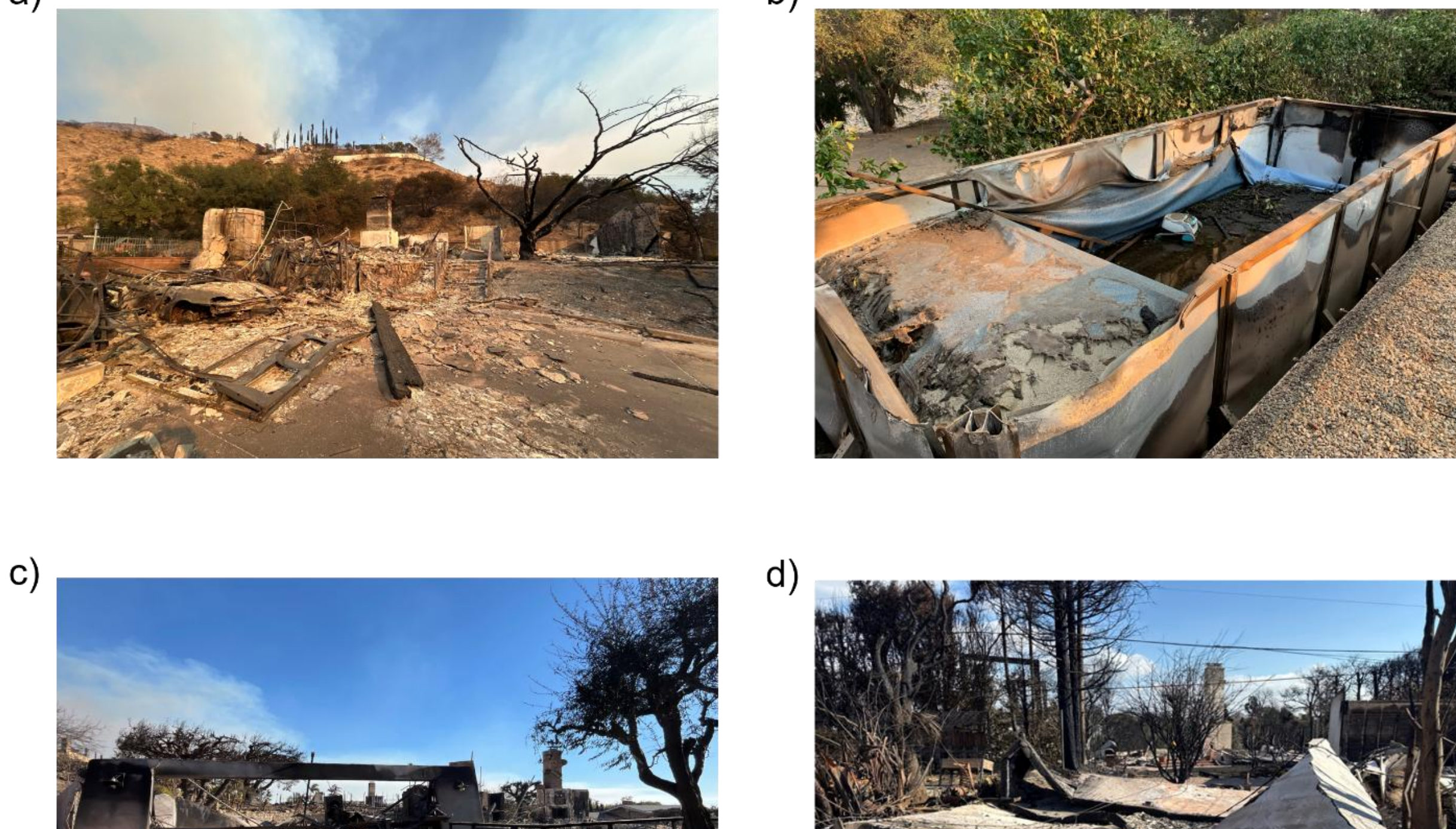

**Figure 18: Each image is an individually *destroyed* (GTL) building in Eaton (a – b) and Palisades (c – d), where Pipeline B – Prompt D misclassified them as *affected.***

This visual analysis highlights the difficulty of classifying subtle damage, such as broken glass, and the importance of prompting methods with clear guidelines for the model to follow. Moreover, it demonstrates the importance of having a workflow where a domain expert can quickly validate the model's claims and identify oversights that would put the public's safety in damage, such as those made in figure 18. The future of this work is promising and shows how MLLM can enhance damage classification, but it is not a sole replacement of damage inspectors. This workflow should be used as a tool to complement and aid the damage inspectors rather than replace their judgement.

## 4. Discussion

Ground-level wildfire damage assessment remains predominantly a manual task, despite the availability of tools that can improve the efficiency and accuracy of this process. This study examines the potential of language models to enhance damage assessment through an end-to-end integrated pipeline and a decoupled vision-text pipeline. To do this, 500 random samples were checked for each study area to ensure the input data was of quality. Once pre-processing was done, the first objective examined the extent to which multi-view images enhance damage classification. The MLLM approach is uniquely suited for this capability, as it processes multiple perspectives as a sequence of embeddings. To establish a baseline, single front-façade images were compared against the multi-view configuration for both pipelines. The single-view assessment yielded insufficient F1 scores for the *affected* category across both Eaton and Palisades datasets, ranging from 0.225 to 0.511. However, the integration of multi-view imagery significantly mitigated these limitations. By processing a list of images simultaneously, both pipelines holistically assess a structure's condition, effectively identifying damage obscured in front-facing views. This approach raised F1 scores for the *affected* category to 0.857 to 0.947, confirming that accurate damage triage requires a comprehensive visual context rather than a single vantage point.

The second objective explored the extent to which prompting methods impact damage classification. By comparing reasoning-based approaches (Prompt C and Prompt D) with zero-shot (Prompt A) and heuristic strategies (Prompt B), the research found that within Pipeline A, the zero-shot and reasoning methods did not differ significantly in accuracy. Instead, the primary trade-off lies in computational efficiency (Prompt A) versus interpretability (Prompt C and Prompt D). Conversely, when examining Pipeline B with Prompt D, the findings suggest that a pure text-based classification approach requires a rigid list rather than open-ended reasoning. This is because the decoupled text model's capabilities rely on structured data ingestion; without direct visual access, it requires structure rather than abstract reasoning to function effectively. While visual classification (Pipeline A) offers superior stability for general detection and the flexibility to include reasoning, a text-based classification approach (Pipeline B) can be beneficial when semantic alignment with policy frameworks dictates the workflow. Unlike vision models that are optimize for abstract pixel patterns, a heuristic text classifier acts as a logic gate, forcing the model to validate specific compliance criteria and strictly abide by policy requirements. Furthermore, this decoupled architecture enables a token-efficient update loop. For example, if new policies emerge that alter the classification criteria for specific damage categories, re-classifying a large visual dataset would typically be computationally expensive. However, a decoupled approach bypasses this cost by re-classifying structures using the cached visual descriptions generated in the initial pass. In this workflow, the only component requiring an update is the prompt classification criterion. This avoids the latency and expense of re-running the vision model on raw imagery, allowing agencies to instantly adapt to new regional taxonomies or evolving hazard definitions with minimal computational overhead. To further explore MLLMs in wildfire damage assessment, a comparison between GPT-4o and Qwen is conducted in supplementary material section *S4*. The research selected the Qwen model due to its synergy with proposed pipelines. Table S4.1 shows that the F1 scores for the *affected* category had a drop off as the Qwen model ranged from 0.739 to 0.862 while the GPT-4o model ranged from 0.845 to 0.947. One contribution to this is that the Qwen model is a 32-billion parameter model while the GPT-4o model is significantly larger. This shows the importance of selecting the optimal open-sourced model when conducting damage assessments.

Conventional deep learning approaches for wildfire damage estimation, such as the ViT model proposed by Luo and Lian (2024), establish a strong supervised baseline for this domain. While their ViT model achieved exceptional performance for the binary extremes of *destroyed* and *no* *damage* (F1 scores of 0.950 and 0.990, respectively), it achieved an F1 score of 0.730 for the *affected* class. In contrast, the multi-view assessment for Pipeline A and Pipeline B outperformed this supervised baseline, achieving *affected* F1 scores ranging from 0.808 to 0.947 (across all prompting strategies for both study areas). In addition to a ViT model, Luo and Lian (2024) tested a zero-shot CLIP and CNN model. These models were only able to obtain precisions ranging from 0.140 to 0.280 for *affected* structures. While these results were obtained on a different dataset, it further highlights the ability of the language models presented in this study. Specifically, the training free approach, especially in the multi-view configuration, outperforms traditional supervised training methods.

This performance comparison reveals a critical operational trade-off. While supervised architectures like CNN and ViT offer distinct advantages in inference speed and lower computational resource—like cost per image after training—they present a significant logistical barrier during the immediate aftermath of a disaster: the absolute requirement for extensive, labeled datasets. This study diverges from the crucial groundwork established by Luo and Lian (2024) as the training phase can be eliminated entirely with MLLMs and LLMs. Moreover, the proposed workflow utilizes a pre-trained multimodal model to enable immediate deployment. This allows practitioners to trade the inference latency of large models for the operational agility of a zero-shot system that requires no lead time for model development. Ultimately, these pipelines are designed to serve as tools for rapid triage rather than definitive certification. The automated classifications enable emergency managers to gain immediate situational awareness and prioritize resources for areas with the most severe damage. By allowing practitioners to fine-tune assessment criteria via prompts—adapting to different hazard types or regional building codes—this methodology offers an agility that static, pre-trained CNN or ViT architectures cannot match.

## 5. Concluding Remarks

The primary contribution of this work is a training-free, multi-view workflow that transforms ground-level images into auditable, indicator-guided damage assessments at the household level without requiring labeled datasets or custom multi-image fusion networks. The framework introduces three specific innovations. First, we operationalize household-level multi-view assessment within a pre-trained vision-language model, enabling comprehensive damage evaluation from multiple perspectives. Second, we implement a structured reasoning layer that extracts binary damage indicators prior to final classification, thereby enhancing both transparency and the ability to modify domain knowledge without retraining. Third, we provide explicit and reproducible design specifications, including the overall framework architecture, cosine-similarity methodology for consolidating *minor*, *affected*, and *major* damage categories, as well as detailed token-level computational and cost analysis. These components combine to create a deployable pipeline that practitioners can customize through simple prompt and indicator modifications rather than model retraining, significantly reducing barriers to rapid post-fire deployment while maintaining scientific rigor.

This framework provides preliminary damage assessment teams with an automated, immediately deployable capability that directly addresses two critical operational challenges identified in the introduction: the resource-intensive and slow nature of manual damage assessments and the limitations of aerial imagery in capturing parcel-level detail due to occlusion and resolution constraints. The system's multi-view approach recovers damage patterns invisible from front façade perspectives alone, dramatically improving *affected*-class classification accuracy from approximately 1–36% with single-views to 77–95% with multiple views across both fire events. This improvement enables rapid identification of properties requiring immediate inspection while reducing costly misclassifications of damaged structures as undamaged.

The system's reliance on pre-trained vision-language models eliminates the need for local training data, allowing immediate deployment following a wildfire event. A comprehensive token-level cost analysis reveals a critical operational trade-off: while the integrated MLLM approach (Pipeline A) offers the lowest initial latency for single-pass assessments, the decoupled Pipeline B enables a workflow that allows practitioners to store visual descriptions and can rapidly re-evaluate damage criteria using a cost-efficient text-based classifier. The flexibility of both pipelines transforms dispersed photographic evidence into actionable, auditable assessments that complement aerial surveys and accelerate equitable resource allocation during the critical initial response phase.

Given the complexities inherent in structural damage assessment, a human-in-the-loop (HITL) approach remains essential for validation and accountability. While the multi-view assessment significantly improved accuracy, the model may still struggle with subtle or ambiguous indicators. As observed in the visual analysis, indicators such as broken glass or the distinction of localized debris can sometimes lead to misclassifications. Therefore, outputs from the model, especially those classified as *affected* or cases where the model confidence is low, must be flagged for manual review by trained personnel. This hybrid approach ensures that the efficiencies of automation are realized without sacrificing the accuracy and nuanced understanding provided by human expertise.

Despite the promising results, this study has some limitations. A primary constraint is the reduced granularity of the damage assessment, necessitated by the aggregation of *minor*, *major*, and *affected* categories due to sample size limitations and high visual similarity between classes. This consolidation obscures nuanced differences that are critical for precise recovery planning. Future research can mitigate this limitation by fine-tuning the pre-trained MLLM model to improve on the categories with similar cosine similarity score. Additionally, fine-tuning the model can improve the MLLM's ability to interpret complex scenarios. For example, the visual analysis revealed challenges in the MLLMs ability to interpret situations such as misclassifying severely damaged but still-standing structures or failing to detect subtle indicators like broken glass. Another limitation is that the current workflow relies exclusively on visual inputs, omitting contextual data that influences structural vulnerability, such as building materials, vegetation proximity, or localized fire intensity. Incorporating these features into the proposed framework offers the potential to enhance the model's classification abilities. These limitations can be addressed by advancing the framework towards a truly multimodal system, integrating ground-level imagery with supplementary data sources like aerial imagery, geospatial data, and tabular structural characteristics. To improve the detection of nuanced damage and enhance classification granularity, advanced techniques such as retrieval-augmented generation should be explored. By grounding the LLMs reasoning in a knowledge base of forensic engineering guidelines and utilizing larger, more diverse datasets, the model's capacity for fine-grained, context-aware damage classification can be significantly enhanced. Additionally, targeted fine-tuning of the MLLM on specialized disaster imagery could improve sensitivity to subtle damage indicators.

Finally, the reliance on ground-level imagery necessitates strict adherence to data privacy protocols. While the images used in this study were sourced from a post-disaster damage inspection database, the broader deployment of GLI collection methods (e.g., crowdsourcing or rapid street-level captures) must adhere to ethical guidelines. Protocols must be established to anonymize any personally identifiable information captured incidentally in the imagery, such as faces, license plates, or house numbers, through blurring or redaction before images are processed and stored. This is vital to protect the privacy of residents during a vulnerable time. Overall, this framework demonstrated the potential that MLLMs and LLMs have in wildfire damage assessment by enhancing the classification of intermediate damage and explores the trade-off of different prompting methods. This provides practitioners with a flexible zero-shot training method that can be immediately deployed in the aftermath of a wildfire event.

## Supplementary Material

### S1. Image analysis

Table S1.1 presents the Pipeline A – Prompt A test with 250, 500, and 750 images in both study areas. Since the results in section 3.6 show that prompting has minimal impact on classification metrics, this pipeline was selected due to its simplicity and token efficiency. The results in this table show no clear drop or gain in the classification metrics based on the number of images since the overall F1 score ranged as follows for (Eaton; Palisades) across each image test per category: *no damage* (0.886 – 0.940; 0.844 – 0.862); *affected* (0.866 – 0.920; 0.818 – 0.857); *destroyed* (0.963 – 0.994; 0.990 – 0.994). This further demonstrates the potential of using a pre-trained MLLM model as the need for large amounts of data is omitted, thus mitigating a limitation when training supervised models.

**Table S1.1: Classification metrics for Pipeline A – Prompt A with varied image samples**

| Metrics | Eaton (250) | Eaton (500) | Eaton (750) | Palisades (250) | Palisades (500) | Palisades (750) |
|---|---|---|---|---|---|---|
| **Accuracy** | 0.916 | 0.942 | 0.936 | 0.880 | 0.900 | 0.901 |
| **Micro-Average F1 Score** | 0.916 | 0.942 | 0.936 | 0.880 | 0.900 | 0.901 |
| **Macro-Average F1 Score** | 0.915 | 0.943 | 0.937 | 0.888 | 0.903 | 0.899 |
| **Cohen's Kappa** | 0.874 | 0.913 | 0.904 | 0.832 | 0.850 | 0.852 |
| **No Damage Category** | | | | | | |
| **Precision** | 0.848 | 0.915 | 0.925 | 0.792 | 0.783 | 0.812 |
| **Recall** | 0.929 | 0.977 | 0.948 | 0.905 | 0.955 | 0.918 |
| **F1 Score** | 0.886 | 0.940 | 0.936 | 0.844 | 0.861 | 0.862 |
| **Affected Category** | | | | | | |
| **Precision** | 0.919 | 0.971 | 0.954 | 0.887 | 0.961 | 0.912 |
| **Recall** | 0.819 | 0.874 | 0.874 | 0.760 | 0.774 | 0.784 |
| **F1 Score** | 0.866 | 0.920 | 0.912 | 0.818 | 0.857 | 0.843 |
| **Destroyed Category** | | | | | | |
| **Precision** | 0.988 | 0.939 | 0.928 | 0.990 | 0.981 | 0.984 |
| **Recall** | 1.00 | 1.00 | 1.00 | 1.00 | 1.00 | 1.00 |
| **F1 Score** | 0.994 | 0.967 | 0.963 | 0.993 | 0.990 | 0.994 |

## S2. An overview of the algorithms for Prompt C and Prompt D

Algorithms 3, 4, and 5 are presented below to show the exact text used for the reasoning prompts and the application of the prompts in the respective pipeline.

**ALGORITHM 3: Pipeline A – Prompt C**

**Input :** Image set I = $\{I_1, I_2, \ldots, I_n\}$, text prompt T∈{Prompt A, C, or D}, vision-language MLLM ∈ {GPT-4o or Qwen}

**Output:** Damage label L ∈ {'Affected', 'No Damage', 'Destroyed'}

**For** each house:

**For** number of self-consistency iterations (i=5):

Prompt T: Prompt C

"Role: You are a DINS (Damage Inspection Specialist). "

"Task: Classify the structure. Be Sensitive to fire damage. "

"Distinguish between 'Cosmetic' (No damage) and 'Actionable' (Affected). "

"### EVALUATION LOGIC ###"

"1. Check for envelope breaches (broken glass, holes). -> Affected.\n"

"2. Check for cosmetic heat damage (blistering paint, soot). -> Affected (High Sensitivity).\n"

"3. Ignore environmental noise (burnt trees, debris piles).\n\n"

"Output JSON:\n"

"{\n"

' "reasoning_trace": "string",\n'

' "final_classification": "No damage" | "Affected" | "Destroyed"\n'

"}"

Output: (Self-Consistency Prediction, Confidence, and reasoning trace)

end

$\{E_1, E_2, E_3\}$ ← Encode up to 3 images for the house in base64 format: $\{I_1, I_2, I_3\}$

Response R ← MLLM(T, $\{E_1, E_2, E_3\}$)

L ← R.label // Model outputs one label based on input images and prompt

**end**

Save

**ALGORITHM 4: Pipeline A – Prompt D**

---

**Input :** Image set I = {$I_1, I_2, \ldots, I_n$}, text prompt T∈{Prompt A, C, or D}, vision-language MLLM ∈ {GPT-4o or Qwen}

**Output:** Damage label L ∈ {'Affected', 'No Damage', 'Destroyed'}

**For** each house:

Prompt T: Prompt D ←

"Role: You are a DINS (Damage Inspection Specialist) creating a dataset. "

"Task: Classify the structure. You must be sensitive to fire damage. "

"Unlike a structural engineer who only cares about safety, a DINS inspector records any heat damage as 'Affected'.\n\n"

"### THE THREE THOUGHT BRANCHES ###"

"1. [Thought A - The 'Pristine' Hypothesis]: Assume the structure is completely untouched (No Damage). To validate this, the paint must be clean, the siding smooth, and the eaves free of soot. **If there is any mark on the building itself, reject this.

"2. [Thought B - The 'Damage' Hypothesis]: Scan for any sign of heat stress (Affected). "

"   - INCLUDES: Soot staining on walls/eaves, blistered/peeling paint, melted gutters, warped vinyl, or charring on attached decks/fences."

"   - INCLUDES: Structural breaches (broken glass, holes)."

"   -> If ANY of these are present on the structure, this hypothesis wins."

"3. [Thought C - The 'Context' Filter]: Critically exclude non-structural noise. Is the 'damage' actually just a burnt bush, a shadow, or other environmental noise? If the house behind the bush is clean, it is No Damage. If the house itself is scorched, it is Affected."

"### EVALUATION & SELECTION ###"

"Compare the visual evidence. Did the fire touch the house? "

"If yes (even just soot/peeling), classify as 'Affected'. "

"Only classify as 'No damage' if the structure is completely clean.\n\n"

"Output your analysis in this JSON format:\n"

"(        '  "thought_A_evidence": "string",\n'

'  "thought_B_evidence": "string",\n'

'  "thought_C_evidence": "string",\n'

'  "winning_hypothesis": "A" | "B" | "C",\n'

'  "final_justification": "string",

"final_classification": "No damage" | "Affected" | "Destroyed")

{ $E_1, E_2, E_3$} ← Encode up to 3 images for the house in base64 format: {$I_1, I_2, I_3$}

Response R ← MLLM(T, { $E_1, E_2, E_3$})

L ← R.label  // Model outputs one label based on input images and prompt

**end**

Save

---

**ALGORITHM 5: Pipeline B – Prompt D**

---

**Input :** Image set $I = \{I_1, I_2, ..., I_n\}$, text prompt $T \in \{\text{Prompt B or C}\}$, vision-language MLLM ∈ {“GPT-4o or Qwen”}

**Output:** Damage label L ∈ {'Affected', 'No Damage', 'Destroyed'}

**For** each house:

Prompt T: Prompt D ←

"Role: You are a DINS (Damage Inspection Specialist) creating a dataset. "

"Task: Classify the structure. You must be sensitive to fire damage. "

"Unlike a structural engineer who only cares about safety, a DINS inspector records any heat damage as 'Affected'.\n\n"

"### THE THREE THOUGHT BRANCHES###"

"1. [Thought A - The 'Pristine' Hypothesis]: Assume the structure is completely untouched (No Damage). To validate this, the paint must be clean, the siding smooth, and the eaves free of soot. **If there is any mark on the building itself, reject this.

"2. [Thought B - The 'Damage' Hypothesis]: Scan for any sign of heat stress (Affected). "

"   - INCLUDES: Soot staining on walls/eaves, blistered/peeling paint, melted gutters, warped vinyl, or charring on attached decks/fences.”

"   - INCLUDES: Structural breaches (broken glass, holes)."

"   -> If ANY of these are present on the structure, this hypothesis wins."

"3. [Thought C - The 'Context' Filter]: Critically exclude non-structural noise. Is the 'damage' actually just a burnt bush, a shadow, or other environmental noise? If the house behind the bush is clean, it is No Damage. If the house itself is scorched, it is Affected."

"### EVALUATION & SELECTION:\n"

"Compare the visual evidence. Did the fire touch the house? "

"If yes (even just soot/peeling), classify as 'Affected'. "

"Only classify as 'No damage' if the structure is completely clean.\n\n"

"Output your analysis in this JSON format:\n"

"(       '  "thought_A_evidence": "string",\n'

'  "thought_B_evidence": "string",\n'

'  "thought_C_evidence": "string",\n'

'  "winning_hypothesis": "A" | "B" | "C",\n'

'  "final_justification": "string",)

LLM prompt $T_2$: Prompt D ←

“You are an expert in post wildfire disaster building damage assessment.

Given the reasoning logic of the final justification for the damage in the JSON, classify the building as “No damage”. “Affected”,  or “ Destroyed”

$\{E_1, E_2, E_3\}$ ← Encode up to 3 images for the house in base64 format: $\{I_1, I_2, I_3\}$

Response R ← MLLM(T, $\{E_1, E_2, E_3\}$)

L ← R.label  // Model outputs one label based on input images and prompt

**end**

Save

---

## S3. McNemar classification tables

To complement the results in table 8, figure S3.1 through figure S3.3 shows contingency matrices of each case for only the *no damage* and *affected* category in each study area. The additional number of households that the multi-view assessment was able to correctly classify for Pipeline A was 116 – 126 (S3.1) and 114 – 125 (S3.2) for Pipeline B. Figure S3.3 shows that Pipeline B was only able to correctly classify 9 – 20 additional households compared to Pipeline A.

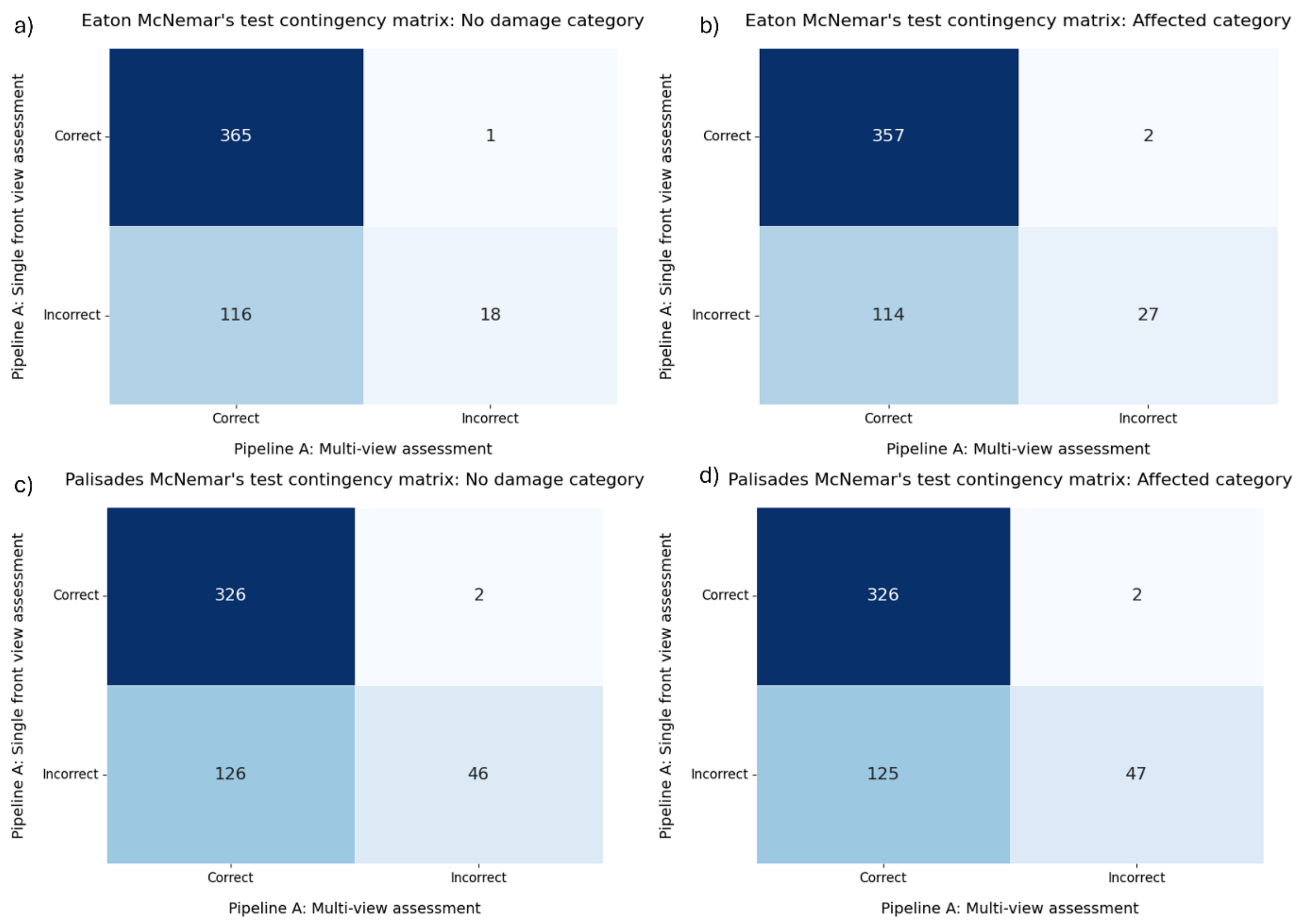

**Figure S3.1: McNemar's test for comparing Pipeline A's single-view (y axis) and multi-view (x axis) classification results. The multi-view assessment had better classifications for *no damage* category in Eaton, 116, and Palisades, by 126. The same can be seen for the *affected* category in Eaton, 114 and Palisades, 125.**

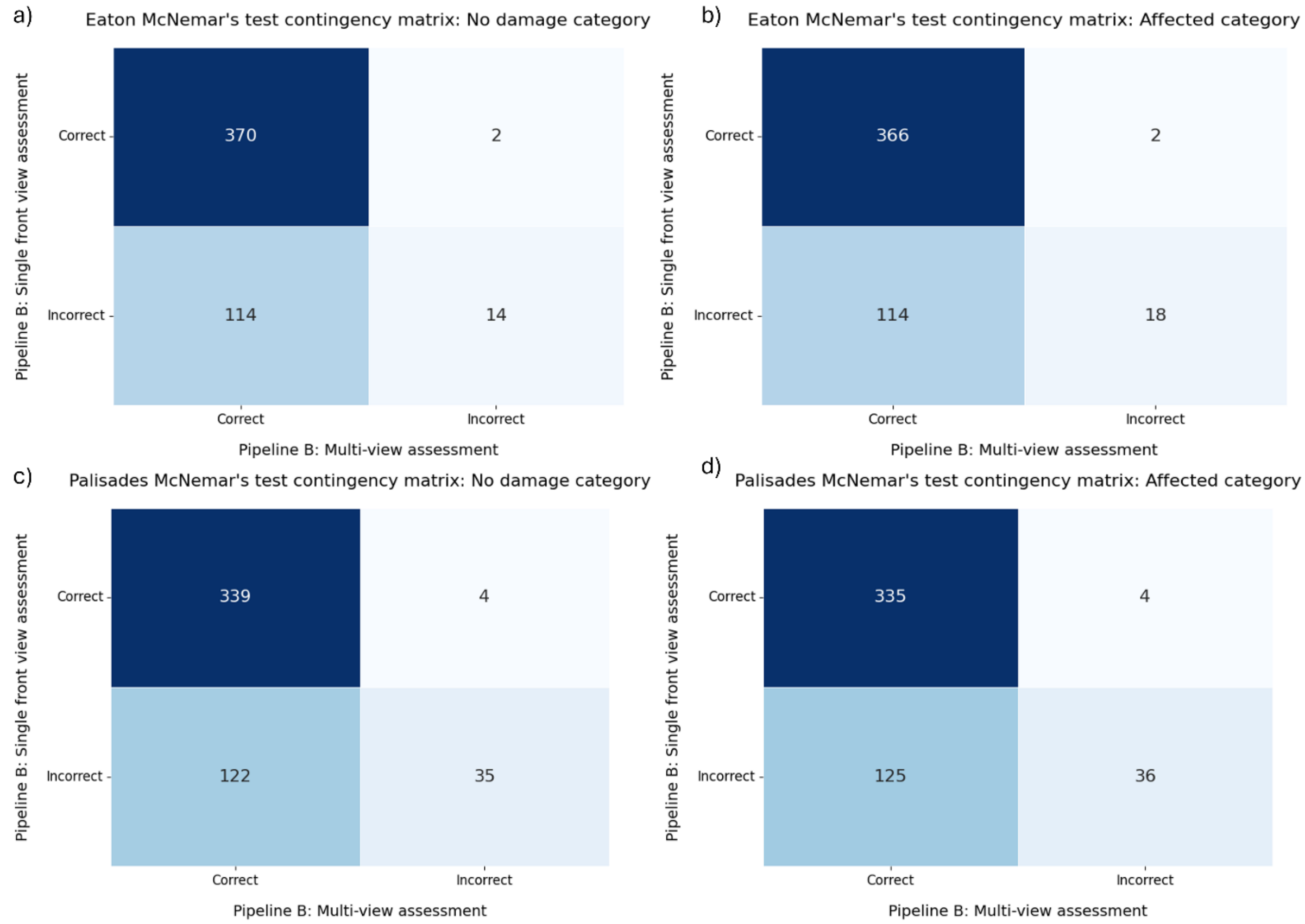

**Figure S3.2: McNemar's test for comparing Pipeline B's single-view (y axis) and multi-view (x axis) classification results. The multi-view assessment had better classifications for *no damage* category in Eaton, by 114 , and Palisades, by 122. The same can be seen for the *affected* category in Eaton, 114 and Palisades, 125.**

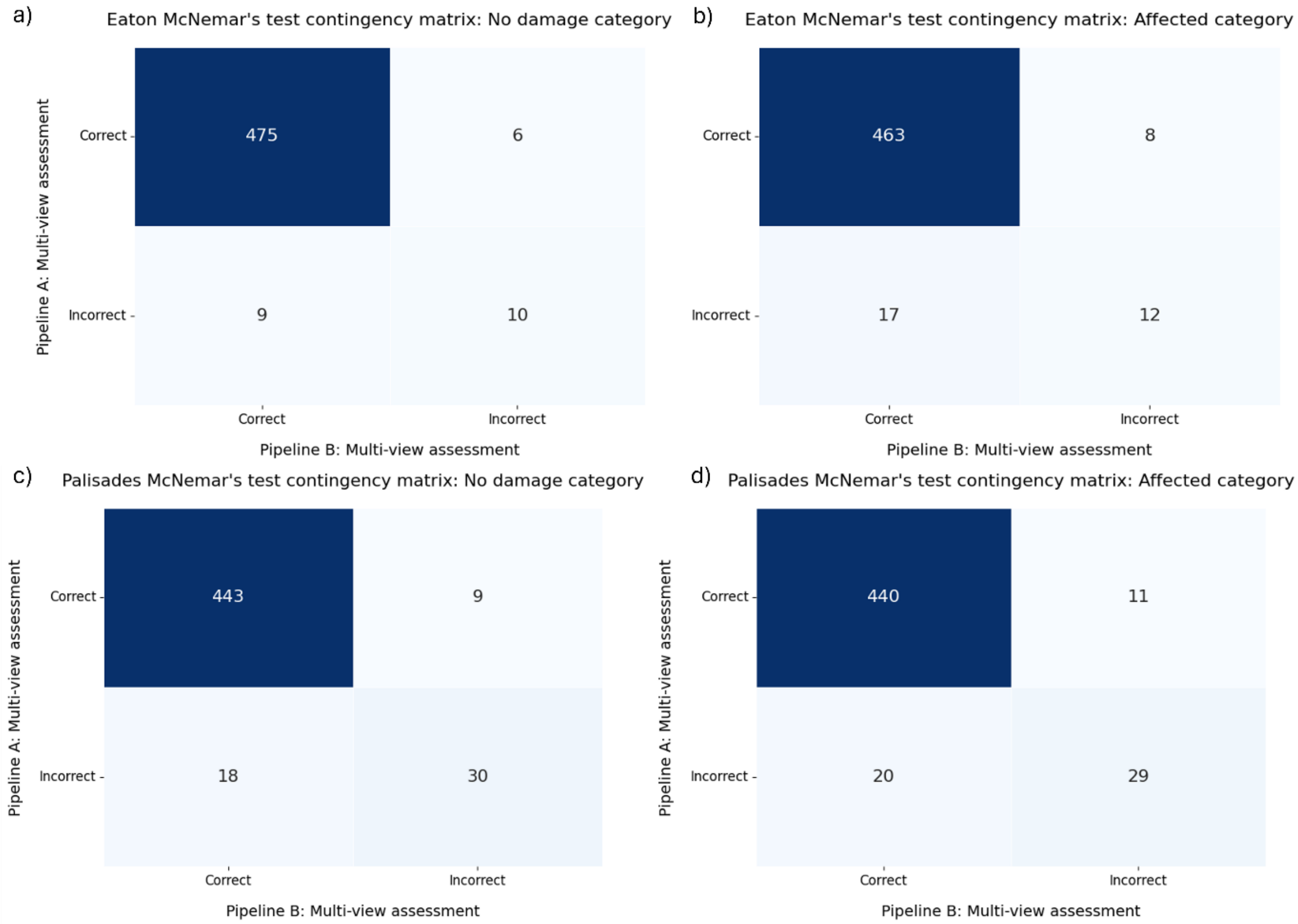

**Figure S3.3: McNemar's test for comparing the multi-view assessment classifications for Pipeline A and Pipeline B. The improvements were not statistically significant as Pipeline B improved S3.3a by nine images, S3.3b by 17 images, S3.3c by 18 images, and S3.3d by 20 images.**

## S4. A multi-view comparison of Qwen2.5-Vision-Language-32-Billion-Instruct

The baseline model was GPT-4o since the time of the writing it was the latest GPT model out. To compare these baseline results to other open-source models, the Qwen2.5-Vision-Language-32-Billion-Instruct (Qwen) model is selected. This open-source model was selected since it could be implemented in Algorithms 1 through 5 easily as the only modification was the API key. For this supplemental experiment, the multi-view configuration was selected. Additionally, the baseline prompts are chosen as well as Pipeline A – Prompt D since it was the most economical reasoning pipeline while maintaining the classifications standards that Pipeline B – Prompt D did not meet.

Table S4.1 shows the Qwen results followed by the GPT-4o (GPT) results for comparison. The Qwen results showed worse classification metrics across all pipelines and prompt pairs for both study areas. Most notably, the F1 scores for the *affected* category had a drop off as the Qwen model ranged from 0.739 to 0.862 while the GPT model ranged from 0.845 to 0.947. One contribution is that the Qwen model is a 32-billion parameter model while the GPT model is significantly larger. Moreover, the way the model “sees” the pixels is different as Qwen uses a dynamic resolution approach that adapts to the aspect ratio of the image. This can sometimes lose details, such as broken glass. Moreover, the GPT model uses a more advance tiling system that breaks the high-resolution image into small sub-grids. This preserves the details of the original image. Pipeline A – Prompt D highlight a further discrepancy, the reasoning of the model. GPT has gone under more reinforcement learning from human feedback that enhances its reasoning, unlike Qwen. These results further highlight another trade-off that can be considered.

**Table S4.1 : Classification metrics for multi-view Images for Qwen and (GPT)**

| **Metrics** | **Eaton Pipeline A-Prompt A Qwen (*GPT*)** | **Eaton Pipeline B-Prompt B Qwen (*GPT*)** | **Eaton Pipeline A-Prompt D Qwen (*GPT*)** | **Palisades Pipeline A-Prompt A Qwen (*GPT*)** | **Palisades Pipeline B-Prompt B Qwen (*GPT*)** | **Palisades Pipeline A-Prompt D Qwen (GPT)** |
|---|---|---|---|---|---|---|
| **Accuracy** | 0.874(*0.942*) | *0.902(0.960)* | 0.652(*0.932*) | 0.828(*0.900*) | 0.842(*0.960*) | 0.732(*0.888*) |
| **Micro-Average F1 Score** | 0.874(*0.942*) | *0.902(0.960)* | 0.652(*0.932*) | 0.828(*0.900*) | 0.842(0.960) | 0.732(*0.888*) |
| **Macro-Average F1 Score** | 0.875(*0.942*) | *0.904(0.961)* | 0.638(*0.933*) | 0.830(*0.903*) | 0.845(*0.961*) | 0.740(*0.892*) |
| **Cohen’s Kappa** | 0.571(*0.913*) | 0.853(*0.940*) | 0.473(*0.897*) | 0.744(*0.850*) | 0.764(0.877) | 0.468(*0.888*) |
| | | | **No Damage Category** | | | |
| **Precision** | 0.861(*0.915*) | 0.851*(0.948)* | 0.660(*0.886*) | 0.688(*0.783*) | 0.702(*0.845*) | 0.620(0.760) |
| **Recall** | 0.877(0.977) | 0.954*(0.948)* | 0.990(*0.955*) | 0.910 (*0.955*) | 0.930(*0.916*) | *0.950(0.961)* |
| **F1 Score** | 0.869(*0.940*) | 0.899*(0.948)* | 0.790(*0.919*) | 0.783(*0.861*) | 0.800(*0.879*) | 0.750(0.845) |
| | | | **Affected Category** | | | |
| **Precision** | 0.885(*0.971*) | 0.927*(0.947)* | 0.550(*0.933*) | 0.902(*0.961*) | 0.907(*0.921*) | 0.710*(0.942)* |
| **Recall** | 0.768(*0.874*) | 0.805*(0.947)* | 0.600*(0.884)* | 0.626(*0.774*) | 0.663(*0.863*) | 0.500(*0.767*) |
| **F1 Score** | 0.823(*0.920*) | 0.862*(0.947)* | 0.572(*0.908*) | 0.739(*0.857*) | 0.766(*0.891*) | 0.590(*0.845*) |
| | | | **Destroyed Category** | | | |
| **Precision** | 0.876(*0.939*) | 0.932*(0.987)* | 1.00(*0.980*) | 0.945(*0.994*) | 0.968(*1.00*) | 0.980(*1.00*) |
| **Recall** | 1.00(*1.00*) | 0.968*(0.987)* | 0.380(*0.954*) | 0.994 (*0.987*) | 0.663(*0.961*) | 0.800(*0.961*) |
| **F1 Score** | 0.934(*0.967*) | 0.949*(0.987)* | 0.550(*0.919*) | 0.968(*0.990*) | 0.971(*0.980*) | 0.88(*0.980*) |

## S5. Reasoning outputs from vision analysis

The tables below are the reasoning outputs for the visual analysis presented in section 3.7 and the summary of these tables can be seen in the descriptions outlined in the aforementioned section.

***Table S5.1 :*** **Prompting outputs for all cases in figure 15, correct classification**

| Image | Pipeline A – Prompt A | Pipeline B – Prompt B | Pipeline A – Prompt C | Pipeline A – Prompt D | Pipeline B – Prompt D |
|---|---|---|---|---|---|
| **Eaton Figure: 15a & 15b** | **Affected** | **Affected**<br>"is the house destroyed": false,<br>"is the structure damaged": true,<br>"is the glass or windows broken": false,<br>"is the furniture burnt": false,<br>"are there burn marks on the structure": true,<br>"is the vegetation around burnt": false | **Affected**<br>The first image shows the front of the house with no visible envelope breaches like broken glass or holes. The second image shows heat damage with blistering paint and soot along the roofline. This indicates cosmetic heat damage. Based on the evaluation logic, the structure is 'Affected' due to the visible heat damage. | **Affected**<br>"thought_A_evidence": "The front view of the house appears clean with no visible damage to the paint or siding.",<br>"thought_B_evidence": "The side view shows clear signs of heat damage, including soot staining and charring on the eaves.",<br>"thought_C_evidence": "The damage is directly on the structure, not obscured by external factors like shadows or vegetation.",<br>"winning_hypothesis": "B",<br>"final_justification": "The presence of soot staining and charring on the eaves indicates heat damage to the structure." | **Affected**<br>"thought_A_evidence": "The front view of the house appears clean with no visible damage to the paint or siding.",<br>"thought_B_evidence": "The second image shows clear signs of heat damage with soot staining and charring on the eaves and roof edge.",<br>"thought_C_evidence": "The damage is directly on the structure itself, not just in the surrounding environment.",<br>"winning_hypothesis": "B",<br>"final_justification": "The presence of soot and charring on the eaves confirms heat damage to the structure, classifying it as 'Affected'." |
| **Eaton Figure: 15c & 15d** | **Affected** | **Affected**<br>"is the house destroyed": false,<br>"is the structure damaged": true,<br>"is the glass or windows broken": false,<br>"is the furniture burnt": false,<br>"are there burn marks on the structure": true,<br>"is the vegetation around burnt": true | **Affected**<br>The images show a structure with intact windows and no visible holes or breaches in the building envelope. However, there is noticeable heat damage, including blistering paint and soot on the window frames and walls. This indicates cosmetic heat damage, classifying the structure as affected. Environmental noise such as burnt trees and debris piles was ignored. | **Affected**<br>"thought_A_evidence": "The structure appears mostly intact with no visible structural breaches.",<br>"thought_B_evidence": "There are signs of soot staining on the walls, blistered paint around the windows, and charred vegetation nearby.",<br>"thought_C_evidence": "The burnt car and debris are not part of the structure itself, but the soot and blistering on the building indicate heat exposure.",<br>"winning_hypothesis": "B",<br>"final_justification": "The presence of soot staining and blistered paint on the building itself confirms heat damage." | **Affected**<br>"thought_A_evidence": "The structure appears mostly intact with no visible structural breaches.",<br>"thought_B_evidence": "There is soot staining on the walls, blistered paint around the windows, and charred vegetation near the structure.",<br>"thought_C_evidence": "Burnt car and debris are present, but they are not part of the structure itself.",<br>"winning_hypothesis": "B",<br>"final_justification": "The presence of soot, blistered paint, and charred vegetation indicates heat damage to the structure, classifying it as 'Affected'." |
| **Palisades Figure: 15e & 15f** | **Affected** | **Affected**<br>"is the house destroyed": false,<br>"is the structure damaged": true,<br>"is the glass or windows broken": false,<br>"is the furniture burnt": false,<br>"are there burn marks on the structure": true,<br>"is the vegetation around burnt": false | **Affected**<br>The inspection of the structure shows no envelope breaches such as broken glass or holes. However, there is visible cosmetic heat damage in the form of charring on the wooden beams, indicating exposure to fire. This is classified as 'Affected' due to the high sensitivity to any signs of fire damage. | **Affected**<br>"thought_A_evidence": "The paint on the walls appears clean and the siding is smooth.",<br>"thought_B_evidence": "There is visible charring on the wooden beams attached to the structure.",<br>"thought_C_evidence": "The charring is directly on the structure, not on surrounding elements.",<br>"winning_hypothesis": "B",<br>"final_justification": "The presence of charring on the wooden beams indicates heat damage.",<br>"final_classification": "Affected" | **Affected**<br>"thought_A_evidence": "The paint on the walls appears clean, and the siding is smooth with no visible marks or soot.",<br>"thought_B_evidence": "There is charring visible on the wooden beams attached to the structure.",<br>"thought_C_evidence": "The charring is directly on the structure, not on any surrounding elements like bushes or shadows.",<br>"winning_hypothesis": "B",<br>"final_justification": "The presence of charring on the wooden beams attached to the structure indicates heat damage, classifying the structure as 'Affected'." |
| **Palisades Figure: 15h & 15g** | **Affected** | **Affected**<br>"is the house destroyed": false,<br>"is the structure damaged": true,<br>"is the glass or windows broken": false,<br>"is the furniture burnt": false,<br>"are there burn marks on the structure": true,<br>"is the vegetation around burnt": false | **Affected**<br>The first image shows a house with no visible fire damage. The structure appears intact with no broken glass or holes. The second image shows significant soot and blistering paint, indicating fire damage. The presence of soot and blistering signifies cosmetic heat damage, making it actionable. | **Affected**<br>"thought_A_evidence": "The first image shows a house with clean paint, smooth siding, and no visible soot on the eaves.",<br>"thought_B_evidence": "The second image shows significant soot staining, blistered paint, and charred materials, indicating heat stress.",<br>"thought_C_evidence": "The damage in the second image is directly on the structure, not just surrounding elements.",<br>"winning_hypothesis": "B",<br>"final_justification": "The presence of soot, blistered paint, and charring on the structure in the second image indicates heat damage.",<br>"final_classification": "Affected" | **Affected**<br>"thought_A_evidence": "The first image shows a house with clean paint, smooth siding, and no visible soot on the eaves.",<br>"thought_B_evidence": "The second image shows significant soot staining, blistered paint, and charred materials, indicating heat stress.",<br>"thought_C_evidence": "The damage in the second image is directly on the structure, not just contextual noise.",<br>"winning_hypothesis": "B",<br>"final_justification": "The second image clearly shows fire damage on the structure itself, with soot and blistered paint, confirming it is 'Affected'." |

### Table S5.2 : Prompting outputs for all cases in figure 16, mixed classification

| Image | Pipeline A – Prompt A | Pipeline B – Prompt B | Pipeline A – Prompt C | Pipeline A – Prompt D | Pipeline B – Prompt D |
|---|---|---|---|---|---|
| **Eaton Figure: 16a & 16b** | **Affected** | **No Damage**<br>"is the house destroyed": false,<br>"is the structure damaged": false,<br>"is the glass or windows broken": false,<br>"is the furniture burnt": false,<br>"are there burn marks on the structure": false,<br>"is the vegetation around burnt": false | **No Damage**<br>The house shows no visible envelope breaches such as broken glass or holes. There is no visible cosmetic heat damage like blistering paint or soot on the structure. The damage to the fence and the presence of debris are considered environmental noise and are ignored. | **Affected**<br>"thought_A_evidence": "The main structure appears clean with no visible soot or charring on the walls or eaves.",<br>"thought_B_evidence": "The fence shows signs of heat damage with warping and charring.",<br>"thought_C_evidence": "The damage is limited to the fence, and the house itself appears unaffected.",<br>"winning_hypothesis": "B",<br>"final_justification": "While the house itself shows no signs of damage, the nearby fence is affected by heat, indicating proximity to fire." | **Affected**<br>"thought_A_evidence": "The front of the house appears clean with no visible soot or damage.",<br>"thought_B_evidence": "The side of the house shows melted and warped vinyl fencing, indicating heat exposure.",<br>"thought_C_evidence": "The damage is directly on the structure's attached fencing, not just nearby vegetation.",<br>"winning_hypothesis": "B",<br>"final_justification": "The presence of melted and warped vinyl fencing attached to the house indicates heat damage, classifying the structure as 'Affected'." |
| **Eaton Figure: 16c & 16d** | **Affected** | **No Damage**<br>"is the house destroyed": false,<br>"is the structure damaged": false,<br>"is the glass or windows broken": false,<br>"is the furniture burnt": false,<br>"are there burn marks on the structure": false,<br>"is the vegetation around burnt": false | **No Damage**<br>The images show a house with no visible envelope breaches such as broken glass or holes. There are no signs of cosmetic heat damage like blistering paint or soot on the structure. The environment depicts burnt trees and debris, but these are considered environmental noise and are not part of the structure evaluation. | **Affected**<br>"thought_A_evidence": "The front of the house appears clean with no visible soot or damage.",<br>"thought_B_evidence": "The back of the house shows signs of damage with charred wood and debris near the deck area.",<br>"thought_C_evidence": "The damage is directly on the structure, not just surrounding elements.",<br>"winning_hypothesis": "B",<br>"final_justification": "The presence of charred wood and debris near the deck indicates heat damage to the structure." | **Affected**<br>"thought_A_evidence": "The paint on the house appears clean, and the siding looks smooth with no visible soot on the eaves from the front view.",<br>"thought_B_evidence": "In the second image, there is visible charring and damage to the wooden deck, indicating heat exposure.",<br>"thought_C_evidence": "The damage is directly on the structure itself, not just in the surrounding environment.",<br>"winning_hypothesis": "B",<br>"final_justification": "The presence of charring on the wooden deck attached to the house confirms heat damage, classifying the structure as 'Affected'." |
| **Palisades Figure: 16e & 16f** | **Affected** | **No Damage**<br>"is the house destroyed": false,<br>"is the structure damaged": false,<br>"is the glass or windows broken": false,<br>"is the furniture burnt": false,<br>"are there burn marks on the structure": false,<br>"is the vegetation around burnt": false | **Affected**<br>1. No envelope breaches such as broken glass or holes are visible.<br>2. Presence of cosmetic heat damage, such as warping or blistering on light fixtures, indicates heat exposure.<br>3. Environmental noise like burnt trees is ignored. | **Affected**<br>"thought_A_evidence": "The structure appears clean with no visible marks, soot, or peeling paint on the main surfaces.",<br>"thought_B_evidence": "The light fixtures show signs of heat stress, with visible warping.",<br>"thought_C_evidence": "There are shadows and natural elements, but the damage to the light fixtures is structural and not contextual noise.",<br>"winning_hypothesis": "B",<br>"final_justification": "The warping of the light fixtures indicates heat damage, classifying the structure as 'Affected'.",<br>"final_classification": "Affected" | **Affected**<br>"thought_A_evidence": "The structure appears clean with no visible soot or charring on the main surfaces.",<br>"thought_B_evidence": "The light fixtures show signs of heat stress, indicating possible exposure to high temperatures.",<br>"thought_C_evidence": "The surrounding area appears unaffected, but the damage to the light fixtures is structural.",<br>"winning_hypothesis": "B",<br>"final_justification": "The presence of heat damage on the light fixtures classifies the structure as 'Affected'." |
| **Palisades Figure: 16h & 16g** | **Affected** | **No Damage**<br>"is the house destroyed": false,<br>"is the structure damaged": false,<br>"is the glass or windows broken": false,<br>"is the furniture burnt": false,<br>"are there burn marks on the structure": false,<br>"is the vegetation around burnt": false | **Affected**<br>The structure does not display any broken windows or significant holes that would indicate an envelope breach. However, there is visible cosmetic heat damage such as blistering paint on the window shutters, suggesting the presence of soot or heat exposure. Given the high sensitivity to cosmetic heat damage, this is classified as 'Affected'. | **Affected**<br>"thought_A_evidence": "The structure has visible soot and discoloration on the shutters and walls.",<br>"thought_B_evidence": "There is soot staining on the shutters and potential heat stress visible on the walls.",<br>"thought_C_evidence": "The damage is on the structure itself, not just surrounding vegetation.",<br>"winning_hypothesis": "B",<br>"final_justification": "The presence of soot and discoloration on the structure indicates heat damage.",<br>"final_classification": "Affected" | **Affected**<br>"thought_A_evidence": "The structure has visible marks and discoloration on the shutters and walls.",<br>"thought_B_evidence": "There is soot staining and possible charring on the shutters and walls, indicating heat stress.",<br>"thought_C_evidence": "The damage is on the structure itself, not just surrounding vegetation.",<br>"winning_hypothesis": "B",<br>"final_justification": "The presence of soot staining and possible charring on the shutters and walls confirms heat damage, classifying the structure as 'Affected'." |

### Table S5.3 : Prompting outputs for all cases in figure 17, all misclassifications

| Image | Pipeline A – Prompt A | Pipeline B – Prompt B | Pipeline A – Prompt C | Pipeline A – Prompt D | Pipeline B – Prompt D |
|---|---|---|---|---|---|
| **Eaton Figure: 17a & 17b** | **No Damage** | **No Damage**<br>"is the house destroyed": false, "is the structure damaged": false, "is the glass or windows broken": false, "is the furniture burnt": false, "are there burn marks on the structure": false, "is the vegetation around burnt": false | **No Damage**<br>The images show a house with no visible broken glass or holes, indicating no envelope breaches. There is no visible cosmetic heat damage such as blistering paint or soot. Environmental elements like burnt trees or debris piles are ignored. | **No Damage**<br>"thought_A_evidence": "The paint appears clean, the siding is smooth, and there is no visible soot on the eaves.", "thought_B_evidence": "No signs of soot staining, blistered paint, melted gutters, warped vinyl, or charring are visible.", "thought_C_evidence": "There are no misleading elements such as burnt bushes or shadows affecting the view of the house.", "winning_hypothesis": "A", "final_justification": "The structure shows no visible signs of fire damage or heat stress." | **No Damage**<br>"thought_A_evidence": "The paint appears clean, the siding is smooth, and there are no visible marks or soot on the eaves.", "thought_B_evidence": "No signs of soot staining, blistered paint, melted gutters, warped vinyl, or structural breaches are visible.", "thought_C_evidence": "There are no misleading elements such as burnt bushes or shadows affecting the perception of the structure.", "winning_hypothesis": "A", "final_justification": "The structure shows no visible signs of heat damage or fire impact. It appears completely untouched and pristine." |
| **Eaton Figure: 17c & 17d** | **No Damage** | **No Damage**<br>"is the house destroyed": false, "is the structure damaged": false, "is the glass or windows broken": false, "is the furniture burnt": false, "are there burn marks on the structure": false, "is the vegetation around burnt": false | **No Damage**<br>Examined images for signs of fire damage. No visible envelope breaches such as broken glass or holes. No signs of cosmetic heat damage like blistering paint or soot. Environmental noise like burnt trees or debris piles is ignored. | **No Damage**<br>"thought_A_evidence": "The paint on the house appears clean, and the siding looks smooth with no visible soot or marks.", "thought_B_evidence": "There are no visible signs of soot staining, blistered paint, melted gutters, warped vinyl, or charring on the structure.", "thought_C_evidence": "The surrounding vegetation appears unaffected, and there are no misleading elements like shadows or tarps affecting the view of the structure.", "winning_hypothesis": "A", "final_justification": "The structure shows no visible signs of heat damage or fire impact. The paint and siding are intact, and there are no signs of soot or charring." | **No Damage**<br>"thought_A_evidence": "The paint appears clean, the siding is smooth, and the eaves are free of soot.", "thought_B_evidence": "No visible signs of heat stress such as soot staining, blistered paint, melted gutters, or structural breaches.", "thought_C_evidence": "No non-structural noise affecting the assessment. The structure itself is clean.", "winning_hypothesis": "A", "final_justification": "The structure shows no signs of fire damage or heat stress. It appears completely untouched." |
| **Palisades Figure: 17e & 17f** | **No Damage** | **No Damage**<br>"is the house destroyed": false, "is the structure damaged": false, "is the glass or windows broken": false, "is the furniture burnt": false, "are there burn marks on the structure": false, "is the vegetation around burnt": false | **No Damage**<br>The first image shows the front of a house with intact windows and no visible signs of damage. The second image shows a close-up of the house exterior with no visible blistering paint or soot. No envelope breaches or cosmetic heat damage are observed. | **No Damage**<br>"thought_A_evidence": "The paint appears clean and the siding looks smooth with no visible marks or damage.", "thought_B_evidence": "There are no signs of soot staining, blistered paint, melted gutters, warped vinyl, or charring visible on the structure.", "thought_C_evidence": "The shadows and objects around the house do not indicate any structural damage.", "winning_hypothesis": "A", "final_justification": "The structure shows no visible signs of fire damage or heat stress. The paint and siding are intact and clean.", "final_classification": "No damage" | **No Damage**<br>"thought_A_evidence": "The paint appears clean, the siding is smooth, and there are no visible marks or soot on the eaves.", "thought_B_evidence": "There is no visible soot staining, blistered paint, melted gutters, warped vinyl, or charring on the structure.", "thought_C_evidence": "There are no misleading elements such as burnt bushes or shadows affecting the view of the structure.", "winning_hypothesis": "A", "final_justification": "The structure shows no signs of heat damage or soot. It appears completely untouched, supporting the 'Pristine' hypothesis." |
| **Palisades Figure: 17h & 17g** | **No Damage** | **No Damage**<br>"is the house destroyed": false, "is the structure damaged": false, "is the glass or windows broken": false, "is the furniture burnt": false, "are there burn marks on the structure": false, "is the vegetation around burnt": false | **No Damage**<br>The images show a residential structure with no visible envelope breaches such as broken glass or holes. There are also no signs of cosmetic heat damage such as blistering paint or soot on the exterior surfaces. The environment appears clean and unaffected by fire damage. Trees and bushes, although leafless, do not show signs of fire damage, which is considered environmental noise and not factored into the structural assessment. | **No Damage**<br>"thought_A_evidence": "The structure appears clean with no visible marks, soot, or damage on the paint, siding, or eaves.", "thought_B_evidence": "No signs of heat stress such as soot staining, blistered paint, melted gutters, or structural breaches are visible.", "thought_C_evidence": "The surrounding vegetation appears dry, but there is no indication of fire damage to the structure itself.", "winning_hypothesis": "A", "final_justification": "The house shows no visible signs of fire damage or heat stress. The structure is clean and intact.", "final_classification": "No damage" | **No Damage**<br>"thought_A_evidence": "The paint appears clean, the siding is smooth, and the eaves are free of soot.", "thought_B_evidence": "No visible signs of heat stress such as soot staining, blistered paint, melted gutters, warped vinyl, or charring.", "thought_C_evidence": "No non-structural noise affecting the assessment. The structure itself appears untouched.", "winning_hypothesis": "A", "final_justification": "The structure shows no signs of fire damage or heat stress. It appears completely clean and untouched." |

**Table S5.4 : Prompting outputs for Pipeline – Prompt D, all destroyed misclassification**

| Image | Pipeline B – Prompt D: Ground truth is destroyed |
|---|---|
| **Eaton Figure 18a** | "thought_A_evidence": "The structure is not pristine; it is heavily damaged with debris and charring visible.",<br><br>"thought_B_evidence": "There are clear signs of heat stress, including charring, soot, and structural collapse.",<br><br>"thought_C_evidence": "The damage is not due to non-structural noise; the structure itself is affected.",<br><br>"winning_hypothesis": "B",<br><br>"final_justification": "The structure shows significant fire damage, including charring and collapse, confirming it is 'Affected'." |
| **Eaton Figure 18b** | "thought_A_evidence": "The structure is not pristine; there are visible signs of damage.",<br><br>"thought_B_evidence": "The structure shows significant signs of heat damage, including soot staining, melted materials, and structural deformation.",<br><br>"thought_C_evidence": "The damage is clearly on the structure itself, not just surrounding elements.",<br><br>"winning_hypothesis": "B",<br><br>"final_justification": "The structure exhibits clear signs of heat damage, including soot and material deformation, confirming it is 'Affected'." |
| **Palisades Figure 18c** | "thought_A_evidence": "The structure is missing significant portions, indicating it is not pristine.",<br><br>"thought_B_evidence": "The structure shows signs of severe damage, including missing walls and roof, and soot staining.",<br><br>"thought_C_evidence": "The damage is clearly on the structure itself, not just surrounding elements.",<br><br>"winning_hypothesis": "B",<br><br>"final_justification": "The structure exhibits clear signs of fire damage, including missing structural elements and soot staining, confirming it is 'Affected'." |
| **Palisades Figure 18d** | "thought_A_evidence": "The structure is not intact; it is completely collapsed.",<br><br>"thought_B_evidence": "The structure shows significant fire damage with collapsed walls and roof, indicating severe heat stress.",<br><br>"thought_C_evidence": "The damage is clearly on the structure itself, not just surrounding vegetation or objects.",<br><br>"winning_hypothesis": "B",<br><br>"final_justification": "The structure is severely damaged by fire, with clear signs of heat stress and collapse, confirming it is 'Affected'." |

## Acknowledgements

The authors would like to thank the members at the Urban Resilience.AI Laboratory for their support, specifically the research assistants Ranvir Patil, Andrew Thomas, Evan Ye, Yahia Ramadan, Yahia Ramadan,  Harsh Dave, and Natalie Mikosz. There was no funding for this research.